\documentclass[review]{elsarticle}

\usepackage{lineno,hyperref}
\modulolinenumbers[5]

\journal{Journal of \LaTeX\ Templates}
\usepackage{siunitx}
 \usepackage[export]{adjustbox}
 
\usepackage{appendix}

\usepackage{longtable}
\usepackage{afterpage}
\usepackage{pdflscape}
\usepackage{enumerate}

\usepackage{amsmath}
\usepackage{amssymb}
\usepackage{multirow}
\usepackage{graphicx}
\usepackage{makecell}
\usepackage{multicol}
\usepackage{graphicx}
\usepackage{array}
\usepackage{graphicx}
\usepackage{multirow}
\usepackage{caption}
\usepackage{makecell}
\usepackage{multicol}
\usepackage{subcaption}
\usepackage{soul}

\usepackage{algorithm}
\usepackage{algpseudocode}

\newcommand\MyBox[2]{
  \fbox{\lower0.75cm
    \vbox to 1.7cm{\vfil
      \hbox to 1.7cm{\hfil\parbox{1.4cm}{#1\\#2}\hfil}
      \vfil}%
  }%
}

\begin{document}
\begin{frontmatter}

\title{Creating awareness about security and safety on highways to mitigate wildlife-vehicle collisions by detecting and recognizing wildlife fences using deep learning and drone technology}

\author[mymainaddress]{Irene Nandutu \corref{mycorrespondingauthor1}}
\author[mymainaddress]{Marcellin Atemkeng\corref{mycorrespondingauthor2}}
\cortext[mycorrespondingauthor1]{irenenanduttu@gmail.com}
\cortext[mycorrespondingauthor2]{m.atemkeng@gmail.com}
\author[mymainaddress]{Patrice Okouma}
\author[mysecondaryaddress]{Nokubonga Mgqatsa}
\author[myadd]{Jean Louis Ebongue Kedieng Fendji}
\author[mymainaddress]{Franklin Tchakounte}


\address[mymainaddress]{Department of Mathematics, Rhodes University, 6139 Makhanda, South Africa}
\address[mysecondaryaddress]{Department of Zoology and Entomology, Rhodes University, 6139 Makhanda, South Africa}
\address[mythirdaddress]{Department of Mathematics and Computer Science, Faculty of Science, University of Ngaoundéré, Cameroon}
\address[myfourthaddress] {Department of SFTI, School of Chemical Engineering and Mineral Industries, University of Ngaoundéré, Cameroon}
\address[myadd]{Department of Computer Engineering, University Institute of Technology, University of Ngaoundéré, Ngaoundéré P.O. Box 454, Cameroon}

\begin{abstract}
In South Africa, it is a common practice for people to leave their vehicles beside the road when traveling long distances   for a short comfort break.  This practice might increase human encounters with wildlife, threatening their security and safety. Here we  intend to create awareness about wildlife fencing, using drone technology and computer vision algorithms to recognize and detect wildlife fences and associated features. We collected data at Amakhala and Lalibela private game reserves in the Eastern Cape, South Africa. We used wildlife electric fence data containing  single and double fences for the classification task. Additionally, we used aerial and still annotated images extracted from the drone and still cameras for the segmentation and detection tasks. The model training results from the drone camera outperformed those from the still camera. Generally, poor model performance is attributed to (1) over-decompression of images and (2) the ability of drone cameras to capture more details on images for the machine learning model to learn as compared to still cameras that capture only the front view of the wildlife fence. We argue that our model can be deployed on client-edge devices to inform people about the presence and significance of wildlife fencing, which minimizes human encounters with wildlife, thereby mitigating wildlife-vehicle collisions. 
\end{abstract}


\end{frontmatter}


\section{Introduction} \label{intro}  
Wildlife fencing systems are popular in Africa and used extensively to minimize wildlife-vehicle collision (WVC) cases~\cite{PEKOR}. The fences are advantageous since they reduce: (1) threats to wildlife from direct human activities (2) conflict between people and wildlife, and (3) zoonotic diseases~\cite{https://doi.org/10.1111/1365-2664.12415}. Fences can also bring about challenges, particularly when they are (1) too low, enabling wildlife to crawl under  easily,  (2) or have loose wires and/or wires spaced too closely together, and (3) are difficult for fleeing animals or birds to see, creating a barrier to wildlife~\cite{hanophy2009fencing}. Barriers of fences to wildlife are (1) limiting foraging opportunities, (2) interfering with connectivity, and (3) failure to easily find new social groups~\cite{PEKOR}. This study captures wildlife fences from South Africa's national routes (N2). The implemented fences are either double or single wildlife fences, and the reserves are fenced following South African laws and regulations~\cite{ZAlawcommission}. The implemented wildlife fences are: (1) single and double fences without electricity, meaning no dangerous animals in the property (i.e. game reserve), (2) the double fences ensure high double protection in case of failure of the inner fence to protect wildlife from escaping. The fence owners have some time to rescue the animal before it runs through the highway fence, and (3) the single and double fences can also be electric, meaning that dangerous animals such as lions (\textit{Panthera leo}) and leopards (\textit{Panthera pardus}) are in the property, and road users' expectations are to take extra precautions while using roads. 
Taking extra precautions when traveling on national roads might lead to a reduction of WVC cases and human deaths. The deaths on highways are worrying in Sub-Sahara Africa~\cite{PEKOR, WendyCollision2015} and South Africa~\cite{AitorAlmeida2018} because of the high numbers of mortality rates of wildlife on highways~\cite{nandutu2022error}. In addition, the deaths soar in South Africa due to tourists' inability to detect wildlife in the neighborhood. This prompts tourists to come out of their vehicles for a short comfort break, making them vulnerable to attacks from dangerous animals. Generally, the high numbers of WVC continue escalating with extreme cases recorded in both developed~\cite{defender,federal2018, Heigl2017} and underdeveloped~\cite{AitorAlmeida2018, WendyCollision2012, WendyCollision2015} countries. These elevated collisions have damaged properties, wildlife, and human lives and are very costly to the economy~\cite{RoadKill}. Identifying ways to minimize these roadkill cases is paramount. Lately, authors are solving this issue by using modeling and prediction~\cite {nandutu2022error} and traditional systems such as overpasses, underpasses, and fencing~\cite{WendyJ2017}. On the other hand, intelligent systems using sensors have also been deployed on highways in developed countries to minimize wildlife mortality rates~\cite{s22072478}. The intelligent systems are; break-the-beam~\cite{MarcelP.Huijser2006}, area cover~\cite{mark2003}, and buried cable~\cite{marcel2012} wildlife detection systems. The intelligent systems detect the presence of wildlife and alert the driver about wildlife on the road before a driver arrives to or near the WVC area.  

Detecting WVC incidences is well attempted in the Northern hemisphere using intelligence systems. In other studies, WVC location detection using classification and clustering algorithms has failed to show acceptable results~\cite{CATLETT201962}. These challenges are solved using time series analysis linear models such as Autoregressive Integrated Moving Average (ARIMA) and autoregressive moving average, which have outperformed state-of-the-art models~\cite{Butt2021}. So far, work on WVC prediction and time series analysis are ongoing~\cite{hp2016roadkill, ozcan2017identifying}, and witnessed in Africa and South Africa~\cite{nandutu2022error, roadkill2015sa}. However, findings show that these ARIMA models have some shortcomings that need to be addressed, and a research gap still exists in this area for future researchers. Some of these shortcomings are the failure to detect a false absence (or presence) of roadkill location and spatial bias in road mortality patterns. Due to weaknesses in previous techniques above, especially those applicable in South Africa~\cite{s22072478} in detecting roadkill areas, we propose and use a novel technique that alerts drivers of the presence of wildlife using a traditional fencing system and computer vision techniques to detect wildlife fences and wildlife fence features. This work is two-fold; (1) we classify and recognize if a wildlife fence is a single or double fence, (2) to minimize noise, we segment wildlife fence insulators as an approach that detects electric wildlife fences.

In this study, we associate areas with wildlife fences and wildlife fence insulators as unsafe for people to get off their vehicles, and drivers should proceed cautiously. Avoiding human encounters with wildlife and WVC incidents helps alert residents and tourists about the safety and security of  a neighborhood, which contributes to improving visitors' safety and minimizing WVC cases. We want to eliminate WVC cases using a cost-effective yet reliable alternative to intelligence systems. We do this by feeding the annotated insulators and their corresponding masks to the U-Net model and fine-tune and concatenate U-Net output prediction. The predictions are then fed into OpenCV to detect insulators on wildlife fences. In addition, single and double fences are input in a classification algorithm to recognize fences. The U-Net, together with OpenCV has characteristics of low cost, high efficiency, and large economic benefits~\cite{liu2020}. U-Net is a popular segmentation algorithm used in medical imaging~\cite{ronneberger2015u}. This architecture is adopted in this work to identify and detect areas where the wildlife fence electrocution and roadkill of wildlife are popular. OpenCV is a cross-platform vision library used to develop real-time computer vision applications. It focuses on image processing, video capture, and analysis, including face and object detection~\cite{bradski2008learning}. 
In addition, classification techniques applied to wildlife fencing are convolution neural networks (CNN) and ResNet. The data collected for this study comprises two datasets, namely, aerial and still images. In this paper, we assessed the prospect of using still and aerial fence images to improve the safety and security of tourists and residents. Our specific objectives were to 1) determine whether wildlife fence and electric fence insulators extracted from still and aerial images could be used as indicators to alert the presence of WVC areas, 2) determine whether the image and associated concatenated mask fed as input to a machine learning algorithm can detect electric fence areas with and without insulators, 3) determine whether the wildlife fence images fed in a classification algorithm can recognize single and double fences and lastly, 4) determine whether OpenCV would detect the electric fence regions with insulators segmented by the machine learning algorithm. Given that wildlife fences and also insulators are unique features on South African highways, we predicted that they would serve as a reliable indicator of informing tourists and citizens about the presence of an animal in the vicinity by sending alerts. This is significant because it detects wildlife fences in real-time, contributing to mitigating WVC and improving tourists' security and safety around game reserves. Additionally, each time a driver detects WVC areas, the client should allow them to save the fence image and global position system (GPS) coordinates. The dataset can further be incorporated into maps or GPS to predict WVC areas and alert drivers on the presence of wildlife on highways via GPS. Alternatively, this algorithm can be deployed in self-driving cars. The images captured by drivers should be used to retrain the model and improve performance via reinforcement learning. Argumentatively, the proposed approach has not been applied before in detecting wildlife fences to mitigate WVC. 

The structure of the work is as follows; Section~\ref{relevance} discusses tourists' security and safety issues as well as strategies to minimize the issues, existing techniques for detecting WVC area and proposes a new technique to detect WVC areas, and finally, wildlife-vehicle collision prevention. Section~\ref{wildlife_fencing} discusses the structure of a wildlife fence and the associated components. Section~\ref{methods} discusses the study's materials and methods. Section~\ref{model_architecture} discusses the deep learning models used and optimization. Section~\ref{evaluation} discusses the evaluation metrics. Section~\ref{results} provides the model experiments and results, and Section~\ref{discussion} discusses the significance of the results, while Section~\ref{conclusion} proposes future works and concludes the study.

\section{Relevance of the work}
\label{relevance}
\subsection{Tourists security and safety}
\label{secuityandsafety}
Tourism is an extremely competitive business, with a core objective of attracting tourists and consequently influencing their tourism destinations. 
A tourism destination is a particular geographic region within which the visitor enjoys various types of travel experiences~\cite{goeldner2007tourism}. In South Africa, the competitive advantage of the tourist sector is popularly backed by science, technology, information, and innovation. The conservationists in this country focus on managing natural resources by creating innovative solutions which contribute to fostering tourist destinations. In this regard, South Africa is a blessed country with a well-connected network of 6.3\% national parks and private nature reserves~\cite{republic1996white}. The latter trend greatly because it meets the demand of the environmentally sensitive visitor. Even though a conducive environment is implemented for tourists in most of these parks, several factors still limit the effectiveness of the tourism industry in South Africa, and these are; (1) fewer funds allocated to the tourism sector, (2) limited integration of local communities and previously neglected groups into tourism, (3) inadequate tourism education, training, and awareness, and (4) tourism security and safety. These challenges discourage potential tourists from visiting South Africa, negatively reducing tourism demand and leading to low tourist destinations. To minimize the challenges, the Western cape government of South Africa implemented a six-leveler strategy for tourism safety, and these are;  (1) a tourism safety forum, which forms a formally structured, multi-stakeholder, and collaborative safety forum, (2) establishing a multi-stakeholder enforcement task team, (3) provision of tourists safety response by providing assistance and supporting tourists while in distress, (4) to identify and develop tourists safety and security technology platform that can act as an enabler to help address and solve core tourists safety risks, (5) to develop a tourism safety comprehensive communication strategy that integrates all important stakeholders, (6) tourism safety interventions and partnerships by identifying synergies that enable funding of cross-platform initiatives and skills developments. Our work implements strategy four of the Western cape government, one of the provinces of South Africa, which proposes to innovate a technology platform that helps address tourists' safety and security risks~\cite{mansfeld2005tourism}. It is worth noting that tourism security is a major constraint to overseas tourism growth. Safety and security considerations are important aspects of increasing tourism destinations in a country. The success or failure of a tourism destination depends on the possibility of providing a safe and secure environment for visitors~\cite{wcg2022}. This study seeks to improve the safety and security of tourists on highways by intervening using artificial intelligence algorithms to minimize wildlife encounters with tourists. Tourists on South African highways need to be made aware of the presence of wildlife fences because some visitors tend to urinate or relax around these demarcated wildlife places enjoying the scenery with no idea that some places have dangerous wildlife. Argumentatively, the machine learning algorithms developed in this study can be incorporated into maps and deployed on car dashboards or driverless cars to automatically detect a wildlife fence and record and save the GPS coordinates and other parameters. The purpose of this work is to inform tourists of the presence of wildlife and wildlife fencing, providing security by minimizing tourists wildlife encounters between wildlife and tourists, which finally leads to the mitigation of WVC.

\subsection{Wildlife roadkill areas detection}
\label{roadkill_hotspots}
Roadkill areas vary depending on the season, spatial and temporal variations, duration of monitoring, and taxa~\cite{oddone2021spatio}.in Researchers in developed economies use intelligent animal detection systems (ADS)~\cite{MarcelP.Huijser2006, marcel2009, WilliamH2019, MarcelHuijser2010, smith2016, shapoval2018, marcel2017, mark2003, MarcelP.Huijser2006, marcel2009, vikhram2017, abir2013, marcel2012, CRISTIANDRUTA2015, CRISTIANDRUTA2020, sharma2017real, rosenband2017inside, sillero2018, Diana2019} to detect roadkill areas. These systems are usually costly, limiting their implementation in Africa since the African governments are poor. Popularly, in time series analysis, the roadkill areas are detected using latitude and longitude coordinates collected by researchers and highway maintenance personnel. However, these detected roadkill areas tend not to align well with sites of most serious animal crossing attempts due to imperfect detections~\cite{hallisey2022estimating}. To address the weakness of imperfect detections and the failure of African governments and organizations to implement a cheaper and novel approach for mitigating WVC. We propose and design an artificial intelligence mitigating strategy using classification and segmentation techniques to detect wildlife fences and their associated features called insulators in WVC locations. The algorithms can further be deployed on client-edge devices to assist drivers in detecting fences when driving the car. The camera can see the wildlife fence image and save the GPS coordinates. Further, such a dataset can be integrated into maps and GPS satellites to predict and visualize future WVC locations to drivers' devices far away before they arrive at a WVC location. Such solutions should be ethically aligned to ensure no harm is caused to animals and humans~\cite{nandutu2021integrating}.

\subsection{Wildlife vehicle collision prevention}
\label{WVCprevention}
The WVC  is a global problem that has escalated in developed and developing countries. United States registers 1,500,000~{WVC} cases, and Canada experiences the death of 4 to 8 {animals} per day~\cite{defender}. In South Africa in 2015, the greater Mapungubwe transfrontier conservation area record shows 991 and 36 WVC on paved and unpaved roads, respectively~\cite{AitorAlmeida2018}. To prevent WVC, mitigation measures are implemented. These are overpasses, underpasses, wildlife fencing, break-the-beam~\cite{MarcelP.Huijser2006, marcel2009, WilliamH2019, MarcelHuijser2010, smith2016}, area cover~\cite{shapoval2018, marcel2017, mark2003, MarcelP.Huijser2006, marcel2009, vikhram2017, abir2013}, buried cable~\cite{marcel2012, CRISTIANDRUTA2015, CRISTIANDRUTA2020}, driver assistance~\cite{sharma2017real}, autonomous vehicles~\cite{rosenband2017inside}, and mobile mapping ADSs~\cite {sillero2018, Diana2019}. Intelligent systems are popular in developed economies, with no attempts made in South Africa.

In this study, we propose and implement a novel mitigation measure that is suitable for the South African economy. It is cheaper and uses existing client-edge devices such as car dashboards and mobile phones. The measures use wildlife fencing, drone technology, classification, and segmentation algorithms to detect wildlife fences and the associated features. The algorithm can be incorporated into maps to detect fences and alert tourists and drivers about wildlife fences on highways. This approach improves the security and safety of tourists and wildlife. In addition, it mitigates WVC.

\section{Wildlife fencing: Single, Double, Electric}
\label{wildlife_fencing}
\begin{figure}
     \centering
     \begin{subfigure}[a]{0.9\textwidth}
         \centering
         \caption{}
         \includegraphics[width=\textwidth,height=6cm]{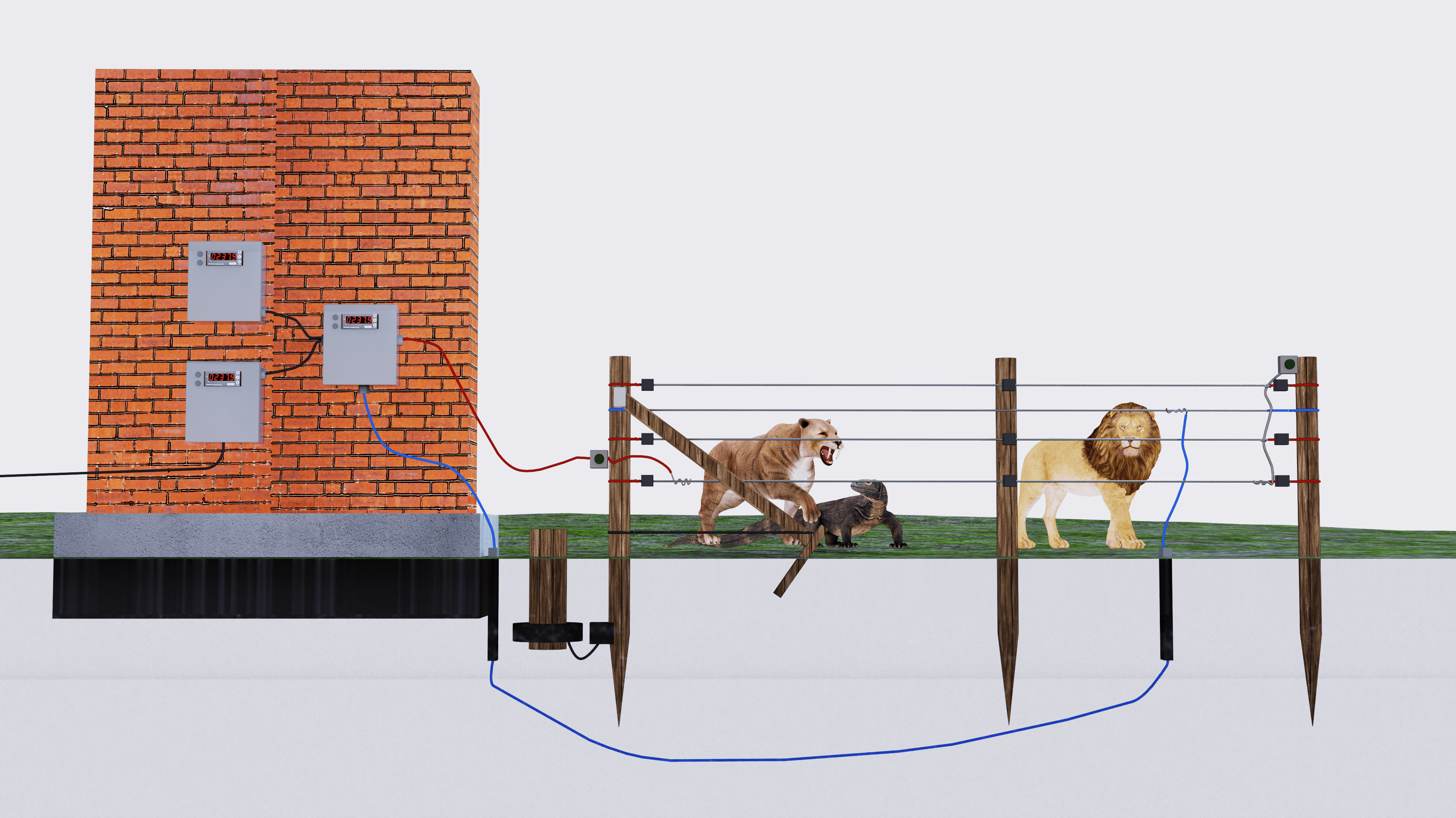}
         
         \label{fig:fence(a)}
     \end{subfigure}
     \hfill
     \begin{subfigure}[b]{0.9\textwidth}
         \centering
          \caption{}
         \includegraphics[width=\textwidth,height=6cm]{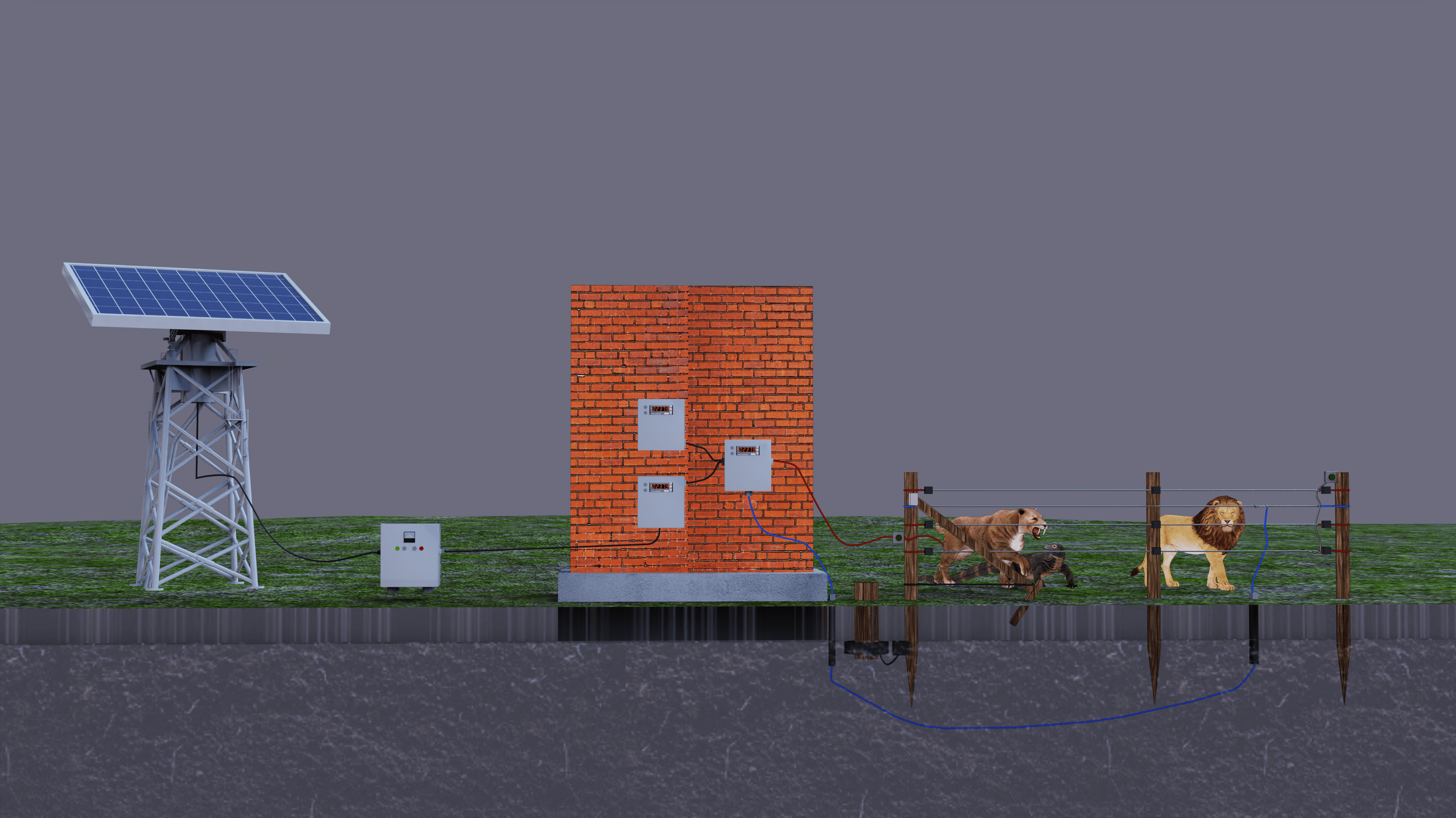}
        
         \label{fig:fence(b)}
     \end{subfigure}
     \hfill
     \begin{subfigure}[c]{0.9\textwidth}
         \centering
           \caption{}
         \includegraphics[width=\textwidth,height=6cm]{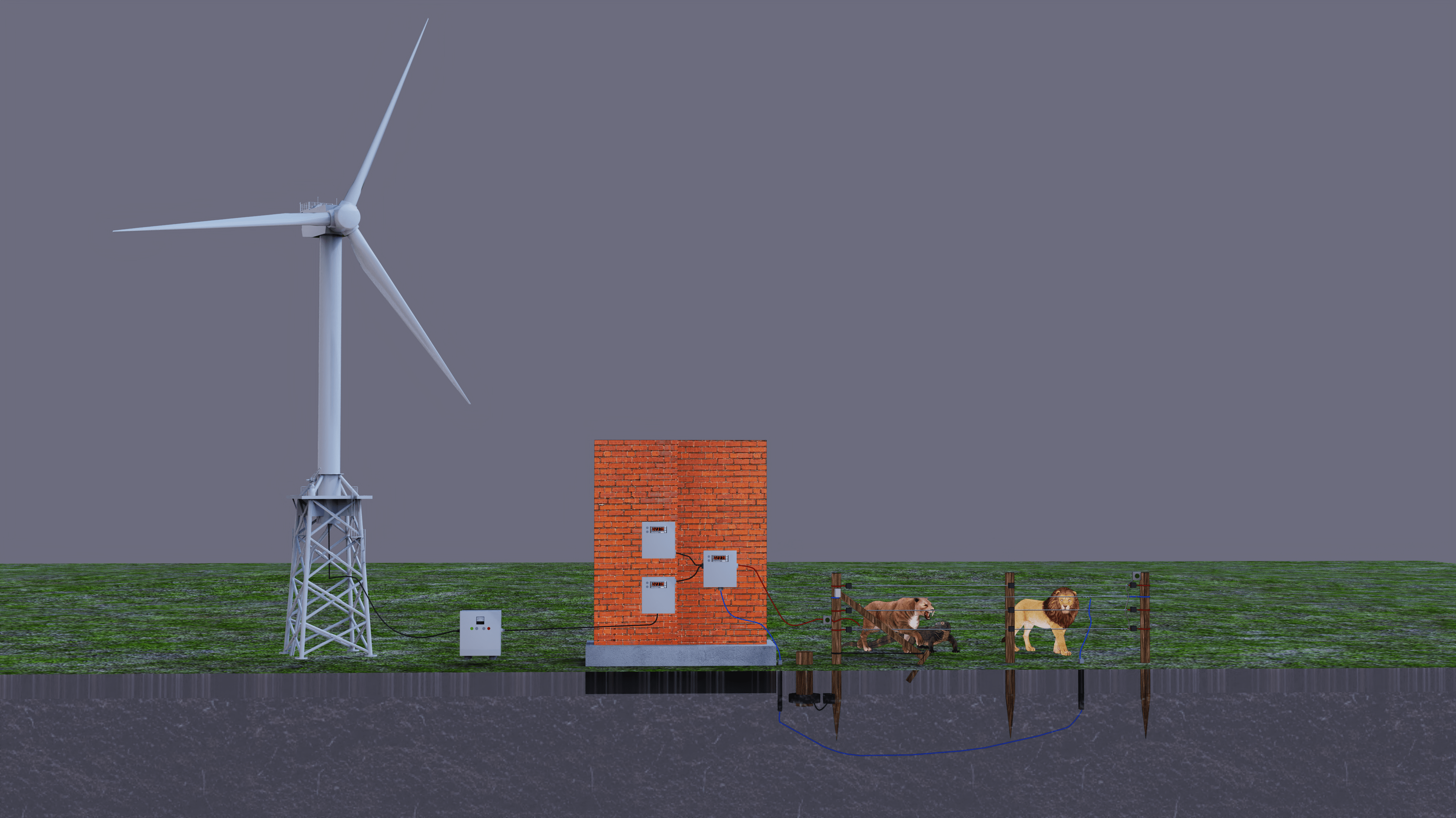}
       
         \label{fig:fence(c)}
     \end{subfigure}
    
        \caption{Structure of wildlife fence showing how electric fence works with components such as in (a) as well as in the solar panel (b) and in wind turbine (c)}
        \label{fig:wildlife_fence}
\end{figure}

Wildlife mortality rates are prevalent in places without traditional mitigation roadkill structures such as the wildlife fences~\cite{roadkill2015sa}. Even in areas with implemented wildlife fencing systems, the wildlife persistently jumps over the fence and collides with cars leading to more cases of WVC~\cite{gagnon2015cost}. Engineers have adopted the installation of electric fences to help mitigate the alarming WVC cases. In designing electric fences, they use a stand-alone fence or in conjunction with other fence types~\cite{ag_fd2022}. In most applications, the fences contain energizers that use batteries recharged from solar panels or wind turbines~\cite{ag_fd2022}. The fences also contain high-tensile electric wire fences with wooden, metal, or plastic posts, as shown in Figure~\ref{fig:wildlife_fence} (a). For an electric wire to work effectively, it requires a set of components such as a \textit{charging system}, \textit{power source}, \textit{regulator}, \textit{energizer}, \textit{earth system},  \textit{conductor},  \textit{reels},  \textit{posts} and \textit{stakes}, and \textit{insulators}. The \textit{plastic standoffs (insulators)} are attached to wooden, metal, or plastic posts~\cite{ag_fd2022}. 

The \textit{charging system} or \textit{power source} is composed of the solar panel and wind turbine as shown in Figure~\ref{fig:wildlife_fence} (c) and Figure~\ref{fig:wildlife_fence} (b), respectively. We note that the wind turbine has a generator and an oscillator that converts the mechanical energy generated by wind to electrical energy. Solar panels generate power by converting sunlight into electricity. Another power source is the battery charged by the solar panel or wind turbine. These two \textit{charging systems} (wind turbine and/or solar panel) produce the voltage and store it in the battery. The stored energy is later used in case of limited energy. The black cable from the wind turbine or solar panel carries or transmits the voltage to the \textit{regulator}. 

The \textit{regulator}, which is a white box on the left in Figure~\ref{fig:wildlife_fence} (b) near the solar panel and Figure~\ref{fig:wildlife_fence} (c) near the wind turbine, controls the voltage needed to kill the animals. For example, assume the wind turbine generates 50 volts or 10 volts. And if the system needs 24 volts, the \textit{regulator} will reduce the volts from 50 to 24 volts or increase the volts from 10 to 24 volts. The volts from the \textit{regulator} are transmitted to the \textit{energizer} (three white boxes) shown in Figures~\ref{fig:wildlife_fence} (a), (b), and (c) through the black cable. The \textit{energizer} has an \textit{earthing system} composed of blue and red cables, which allows the transfer of electronics to and from the fence. The \textit{energizer} achieves the former by pushing an electric pulse through the red wire to the fence; the pulse travels through the conductors and pressurizes the fence with excess electrons. That pressure is measured in volts. When an animal touches the fence or wire, excess electrons enter it and travel through the animal to the soil. After exiting the animal, the pressurized electrons travel through the soil's moisture back to the energizer's \textit{ground rod} (blue cable). The electrons enter the energizer via the \textit{ground rod} (blue cable). The amount of electrons that return is equal to the shock effect. An animal standing on the ground and touching the electrified wire completes the circuit and receives intermittent but regular shocks to deter it. The pulsed nature of the electricity enables animals to move away from the fence, preventing electrocution.  

\textit{Conductors} or wires are equally an essential component of the fence as they conduct electricity along the fence line. Some examples of a conductor are steel wires. Also, \textit{reels} significantly reduce the time it takes to dismantle a fence. A \textit{reel post} allows the fence wires to be independently tensioned and keeps the wire fence taut. Permanent electric fence systems commonly use wooden \textit{posts} with \textit{insulators} attached to them. \textit{Insulators} prevent the conductor or wire from touching the \textit{posts} or \textit{stakes}, which could lead to power leakage~\cite{farm_fence2022}. A good \textit{insulator} is vital in maintaining the fence's performance. In this study, we use wildlife fences and insulators, a feature of a wildlife electric fence, to represent WVC areas where there is a concentration of wildlife. We feed the annotated insulators and their corresponding masks to the U-Net model, fine-tune and concatenate U-Net output prediction. Next, the predictions are fed into OpenCV to detect insulators on wildlife fences.

\section{Data collection and annotation}
\label{methods}
\subsection{Study area}
\begin{figure}
\centering
\includegraphics[width=\textwidth,height=8cm]{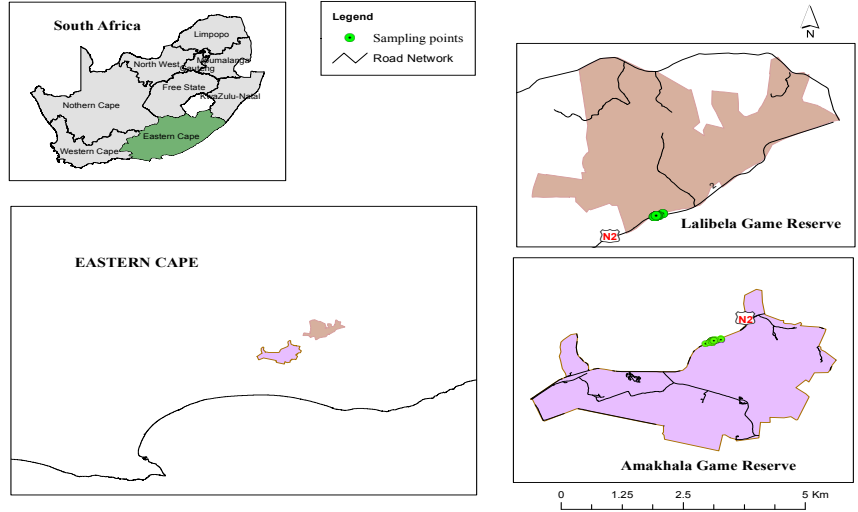}
\caption{The Amakhala and Lalibela game reserves, Eastern Cape Province, South Africa, showing sampling points in green.}
\label{fig:location}
\end{figure}

We took the images used in this study at Amakhala and Lalibela Private Game Reserves in the Eastern Cape province of South Africa (Figure~\ref{fig:location}). The game reserves are located on the highways on a flat and low terrain surrounded by green vegetation. We captured images of the experimental site using a drone and still cameras. To capture aerial images, a DJI Phantom 4 Pro UAV drone is used and Nikon Z6 Mk 2 still camera helped in capturing still images. The Nikon Z6 Mk 2 contains a 24.5MP FX-Format BSI CMOS Sensor, Actual: 25.28 Megapixel, Effective: 24.5 Megapixel  $(6000 \times 4000)$ 
ISO 100-51200, In-body 5-axis Vibration Reduction Sensor incorporates a 273-point phase-detection AF system, Mechanical Focal Plane Shutter, and Electronic Shutter, and the Image File Format is RAW. The UAV has a 1" 20MP CMOS Sensor, Gimbal-Stabilized 4K60 20MP Imaging, Ocusync Transmission, Flight Autonomy with Redundant Sensors, five directions of obstacle sensing, and four directions of obstacle avoidance, Maximum Flight Time of 30 Minutes, Hovering Accuracy ±0.98' / 0.3 m Horizontal with Vision Positioning
±0.33' / 0.1 m Vertical with Vision Positioning and still Image Support – DNG. The highest flying height we went with the drone was 204 meter.

\subsection{Data Acquisition}
\label{dataAcquisition}
The data collection comprised three researchers, one professional photographer, and one UAV pilot. The photographers captured scenes of wildlife fences based on the research objective. Researchers stated the key features to focus on while photographing the wildlife fences. The data collected contains 52 and 159 images from the drone and still cameras, respectively, as shown in Table~\ref{tbl:raw_data}. For both devices, we stored the photos in jpeg format. Drone images have a height and width of 5904 and 3933 pixels, respectively. And still, images have a height and width of 5904 and (3928 or 3936) pixels, respectively. Both of very high resolution as observed.

\begin{table}
\caption{RAW data collected on wildlife N2 fence and data augmented and used for training machine learning models}
\begin{center}
\begin{tabular}{llllll}
\multicolumn{1}{c}{\bf Dataset name}&\multicolumn{1}{c}{\bf Camera type}&\multicolumn{1}{c}{\bf Fence type} &\multicolumn{1}{c}{\bf Images} &\multicolumn{1}{c}{\bf Description}
\\ 
\hline
\hline
Drone dataset & Drone & \makecell{Single  fence}&  26 &\makecell{Images of wildlife N2\\ single fences from drone camera } \\
\cline{3-5}
 & &\makecell{Double fence}  &26&\makecell{Images of wildlife N2 \\double fences  from drone camera}  \\
 
\hline
Still dataset & Standalone & \makecell{Single fence}&  79 &\makecell{Images of wildlife single \\fences  from standalone camera} \\
\cline{3-5}
& & \makecell{Double fence} &80& \makecell{Images of wildlife double \\fences from standalone camera}\\
\hline

Total & & &211&\\
\hline
\label{tbl:raw_data}
\end{tabular}
\end{center}
\end{table}

\subsection{Data preprocessing and annotation}
\label{processandannotate}
The images were in image file format – RAW. This file type is uncompressed and commonly captured by digital cameras and sensors. We processed the images from RAW files to JPEG file format using Adobe Photoshop to convert the format and filter the converted images. 

\begin{figure}
\centering
\includegraphics[width=\textwidth,height=8cm]{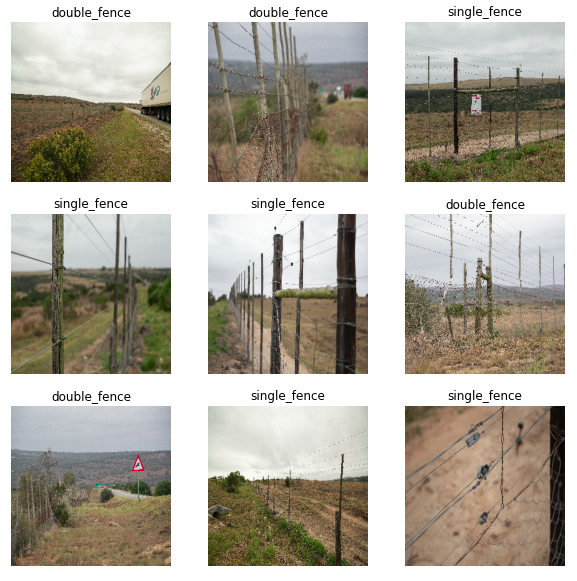}
\caption{Random samples from wildlife fences datasets for Classification Task}
\label{fig:sample_images}
\end{figure}

The original dataset consists of 211 images collected using a standalone camera and drone. For the classification task, the images are compressed to equal dimensions of a width of 512 pixels and a height of 512 pixels. For the two datasets, images from a drone and still cameras are saved in separate folders. The still images contained only images that used a standalone camera; the drone captured aerial images of wildlife fences and scenes from the sky. The drone camera images used in this study are 26 from the single fence and 26 from the double fence. As well as 159 images from a standalone camera containing 79 single and 80 double fence images as shown in Table~\ref{tbl:raw_data}. In Figure~\ref{fig:sample_images}, we visualize a random sample of images from our dataset and deploy it on mendeley~\cite{datainbrief}. Because of few samples, there was a need to increase the size of the training images to improve model accuracy. Data augmentation techniques are applied to expand on the sample of the training dataset. The technique makes variations to the training sample, avoids overfitting, and increases robustness~\citep{shorten2019survey}. Data augmentation transformations applied in this study include horizontal flips, vertical flips, random crops, and a small gaussian blur of about 50\% of all images with random sigma between 0 and 0.5, strengthening or weakening the contrast in each image. We also added some gaussian noise per pixel and channel by sampling only 50\% of all images; this can change not only the color but also the brightness of the pixels. We further made some images brighter and some darker by sampling 20\% of the images of all cases. We sampled the multiplier once per channel, which ended up changing the color of the images. Finally, we apply affine transformations to each image by scaling, zooming, translating, moving, rotating, and shearing them. All the augmentations performed are applied randomly.

\begin{figure}
		\centering
		\includegraphics[width=\textwidth]{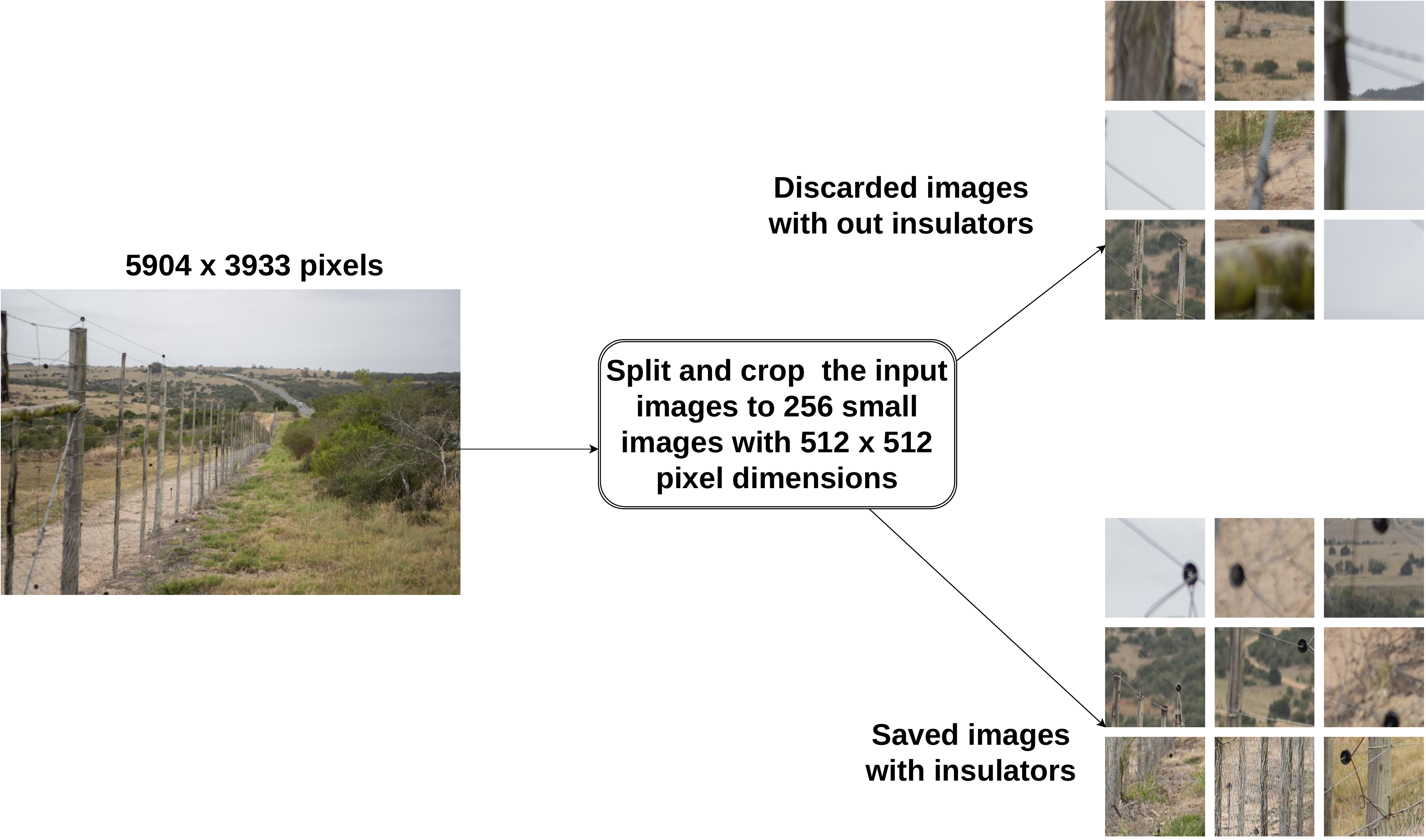}
		\caption{Split the original image into smaller images of $512 \times 512$ pixels. Discard images with no insulators and retain images with insulators for further processing}		\label{fig:slicing}
	\end{figure}

For the segmentation tasks, the original JPEG images have high-resolution pixels as described in Section~\ref{dataAcquisition}. In our observations, the wildlife fence images contain a lot of noise, such as grass and trees, compared to the fence itself. Besides, these images are enormous for machine learning models to process. To minimize the noise, we cropped out noisy parts of the images to $512 \times 512$ pixels. 
The sliced images obtained contain both images with no insulators (noise) and images with insulators. We discarded images with only noise and retained images with insulators, as shown in Figure~\ref{fig:slicing}, where we input the original images and output the cropped images. For each original image, we reshape it to obtain an equal dimension image size for easy cropping into smaller images, which are easily processed by the U-Net algorithm. We then sort the photos into positive and negative regions. Further, we keep the positive regions for further processing and discard the negative regions.

\begin{figure}
		\centering
		\includegraphics[width=\textwidth]{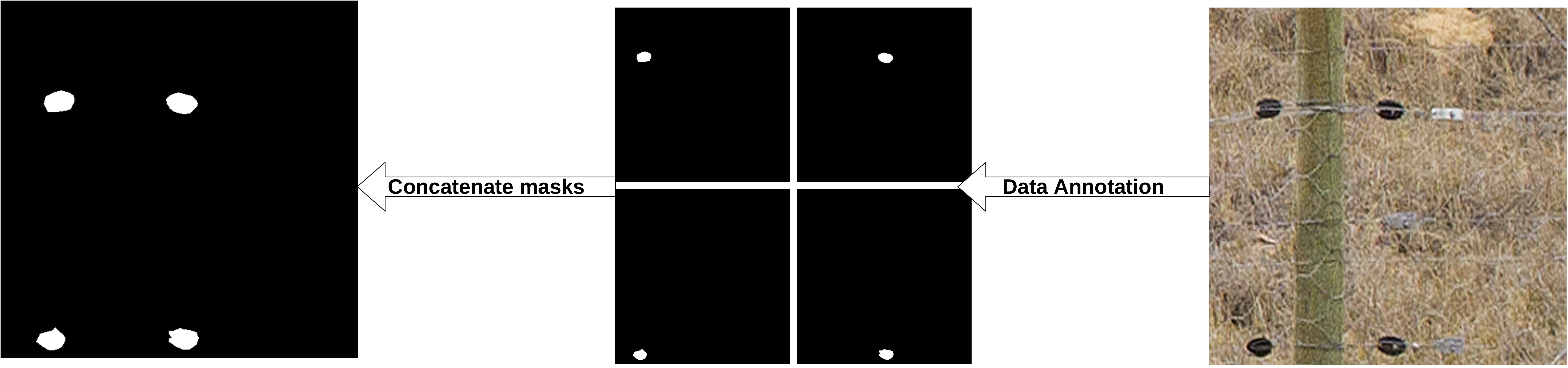}
		\caption{Input the sliced $512 \times 512$ pixel image for annotation to generate individual masks for each insulator in the image, later we combined the individual masks to resemble the original image}		\label{fig:concatenate}
	\end{figure}

To build the ground truth mask of the insulators, we use the Visual Geometry Group (VGG) Image Annotator (VIA)~\cite{dutta2019via}. The VIA is used to define regions and textual descriptions of the regions. The annotated images are saved in a .json extension file containing the coordinates of the insulators. In this task, each sliced image of a wildlife fence insulator is annotated to label insulators as positive regions. The rest of the objects in the image are labeled as negative regions. The positive regions are identified as 1, and the negative region is identified as 0, which gives us a gray-scaled image as an output or binary mask. The results from the annotations are image masks for each insulator found on the wildlife fence. Since VIA output is smaller images with a single insulator, to reconstruct a larger image containing all insulators in an image, we concatenate them, as shown in Figure~\ref{fig:concatenate} and Algorithm \ref{alg:2}. 

The generated pairs of $512 \times 512 $ pixel image masks with associated original images, as shown in Figure~\ref{fig:concatenate} are fed into a neural network for model training and fitting to learn to find the positive regions (insulators) on the wildlife fences images. The annotated images are shown in Table~\ref{tbl:processed_data}. It includes 447 and 686 aerial and still images, respectively, extracted from single and double fences. These are split into the train, validation, and test sets. The still dataset contains 80\% samples for training, 10\% for validation, and 10\% for the test. In addition, drone datasets have 80\% images for training, 10\% for validation, and 10\% for the test.

\begin{figure}
		\centering
		\includegraphics[width=\textwidth]{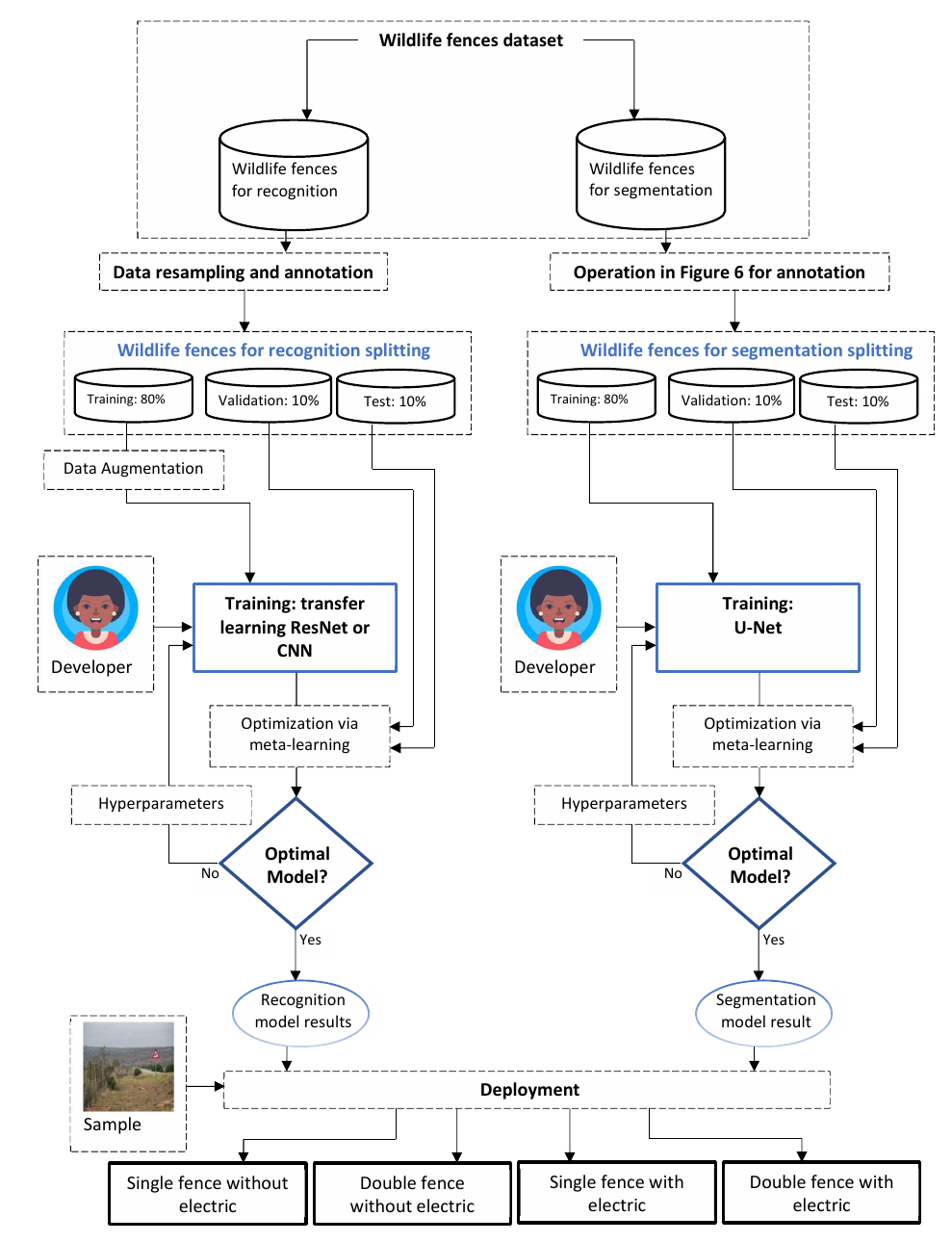}
		\caption{Proposed methodology: The process shows the proposed pipeline  from the recognition and detection implementation to the test from  field data.}		\label{fig:pipeline}
	\end{figure}

Figure~\ref{fig:pipeline} shows all the processes described above, i.e. from fieldwork for data collection, to the implementation of the recognition and detection models then the testing from field data.

\begin{algorithm}
\caption{An algorithm for image annotation and mask concatenation}\label{alg:2}
\begin{algorithmic}
\State Cropped mask $\gets C_m$  
\State Combined cropped mask $\gets CC_m$ 
\State positive regions  $\gets P_r$ 
\renewcommand{\algorithmicrequire}{\textbf{Input:}} 
\renewcommand{\algorithmicensure}{\textbf{Output:}} 
\Require $P_r$
\Ensure $C_m, CC_m$ 
\For{ each $P_r \in \{1,2,3,4,5,6,....n\}$ } 
\If{$P_r$ == $ 512 \times 512 $ pixels} 
    \State Annotate  $P_r \gets C_m, CC_m$ 
    \Else
    \If{$C_m$ == 1} 
    \State width, height = $C_m$.size 
    \State $C_m$ = width, height 
    \State $C_m$ == $P_r$ 
        \State return $C_m$ 
        \Else
         \If{$C_m > 1$} 
         \State $total\_width = \sum(width)$ 
        \State $total\_height = \max(height)$ 
         \State $CC_m$ = $C_m.new(total\_width, total\_height)$ 
         \State $CC_m$ == $P_r$ 
         \State return $CC_m$ 
    \EndIf
    \EndIf
      \EndIf
\EndFor
\end{algorithmic}
\end{algorithm}

\section{Deep learning and optimization}
\label{model_architecture}
\subsection{Convolutional Neural Networks}

Convolutional Neural Networks (CNN) consist of input and output layers. In between these two layers, there are hidden convolutional layers, pooling layers, and a fully connected layer. The layers are made up of neurons that contain learnable weights and biases. Each neuron takes in several inputs, computes a weighted sum by passing it through an activation function, and returns an output~\citep{albawi2017understanding}. CNN's are used to classify images by learning intrinsic image features, clustering these images by similarities, and finally, performing object recognition. A CNN takes in images as input and allocates importance by applying the learnable parameters (weights and biases) to several features in the image, and by doing so, it can differentiate one image from the other~\cite {o2015introduction}.

\begin{table}
\caption{Data reconstructed after slicing, containing images with insulators and annotated masks fed as input to the U-Net model}
\begin{center}
\begin{tabular}{lll}
\multicolumn{1}{c}{\bf Fence type} &\multicolumn{1}{c}{\bf Images} &\multicolumn{1}{c}{\bf Description}
\\ 
\hline
\hline
Drone &  447 & Wildlife fences insulator images and masks from drone camera \\
Still   &686& Wildlife fences insulator images and masks from still camera \\
\hline
Total & 1033& \\
\hline
\label{tbl:processed_data}
\end{tabular}
\end{center}
\end{table}


\subsection{Transfer Learning ResNets}
This study leverages transfer learning using the wildlife fences dataset. An approach is known for developing novel applications in image classification, localization, detection, and segmentation tasks. To improve the performance of CNNs, a transfer learning approach was introduced. Transfer learning is a technique in machine learning where a model is pre-trained on a given dataset and then re-used as a building block in a related task~\cite {saha2018transfer}. The reasoning behind transfer learning is to use a model trained on a certain task to improve learning in a related task of interest. 
Transfer learning has been successful in many machine learning applications, including image classification~\citep{zhu2011heterogeneous}, image localization and detection~\citep{shin2016deep} as well as image segmentation~\citep{van2014transfer}, and has led to the development of CNN models that offer a much greater performance ~\citep{saha2018transfer,weiss2016survey}. Examples of large transfer learning models include VGG, GoogleNet (Inception), and Residual Network (ResNet). In this study, we adopted ResNet50 architecture because of its power of skip connections that eliminates the vanishing gradient problem. We draw upon these works during model development and hyperparameter tuning and the overall fit of the training data to each binary classification model, as it suggests that model overfitting may be perilous to transferring learning to validation and testing sets. The ResNet is developed to overcome some challenges experienced when using the initially created transfer learning models. These common challenges include problems with degradation, network optimization, and vanishing or exploding gradient issues. ResNet50 model uses the concept of shortcut or skipping in the CNN to carry out the task~\citep{he2016deep,he2016identity}. ResNet has shown great performance when applied to computer vision tasks.


\subsection{U-Net for segmentation}
\begin{figure}
		\centering
		\includegraphics[width=\textwidth,height=10cm]{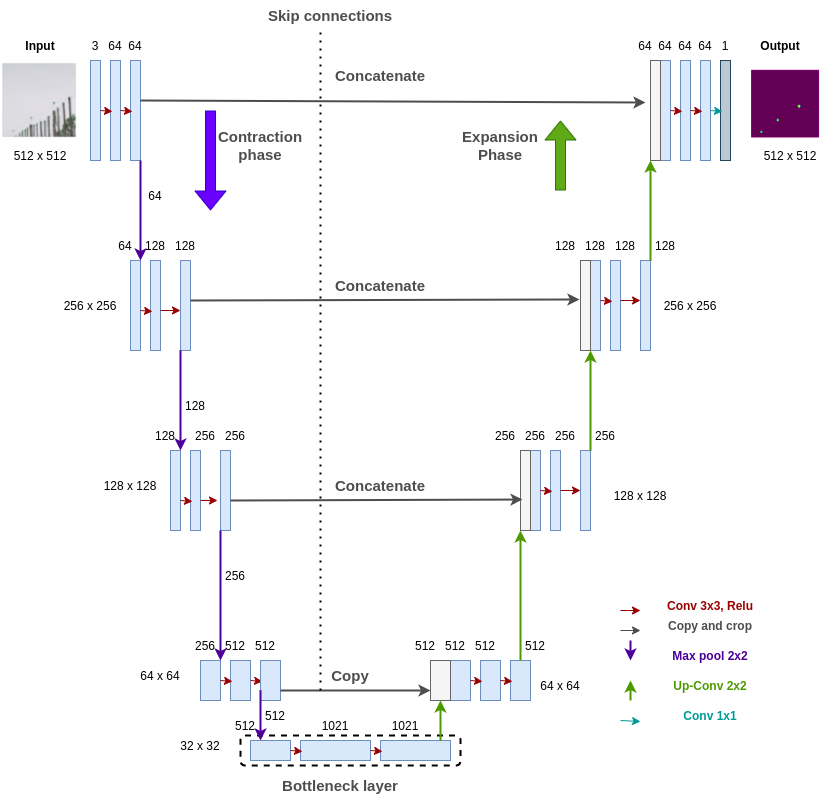}
		\caption{U-net architecture with the arrows represent the different operations, the blue boxes represent the feature map at each layer, and the light gray boxes represent the cropped feature maps from the contracting path and dark gray represents the output image channels}		\label{fig:unet_architecture}
	\end{figure}

The U-Net network architecture consists of an encoder (downsampler) and decoder (upsampler) with a bottleneck connecting the encoder and decoder~\cite{ronneberger2015u}. The skip connections concatenate encoder outputs to each stage of the decoder. The decoder or contracting path contains a convolutional network with alternating $3 \times 3$ convolutions followed by ReLU activation unit as shown in Equation~\ref{conv_relu} and max pooling operations as shown in Equation~\ref{max_pool}. Each stage has an increasing number of filters, and the dimensionality of the features is reduced because of the pooling layer. Later, more features are extracted by the bottleneck. Finally, the expansion path or the decoder increases the resolution of the output by upsampling the features using $2 \times 2$ up-convolution back to the original image size as shown in Equation~\ref{up_sampling}. To localize the object of interest, upsampled features, the expansive path combines them with high-resolution features from the contracting path via skip connections by cropping and concatenating the features. At each upsampling level, we take the output of the corresponding encoder or contracting path and concatenate it before feeding it to the next decoder or expansion path. We use two successive $3 \times 3$ convolutions and ReLU activation during upsampling. At the final stage, we apply an additional $1 \times 1$ convolution to reduce the feature map to the required number of channels and produce the segmented image as shown in Figure~\ref{fig:unet_architecture}. During the expansion path, the cropping of images is vital since pixel features in the edges have the least amount of contextual information and therefore need to be discarded, which results in a network resembling a u-shape. More importantly, it propagates contextual information along the network, allowing it to segment objects in an area using context from a larger overlapping area. The model output is a pixel-by-pixel mask showing the class of each pixel.

\begin{align}  
\mathbf{b}_{x, y, l} &= \mathrm{ReLu}\bigg(\sum_{i,j\in \{-1,0,1\}}\sum_{k}^{}\mathbf{w}_{i, j, k, l}\cdot \mathbf{a}_{x + i, y + j, k} + \mathbf{c}_{l}\bigg)
\label{conv_relu}\\ 
\mathbf{b}_{x, y, k} &= \max_{i, j\in \{0,1\}}\bigg( \mathbf{a}_{2x + i, 2y + j, k}\bigg)\label{max_pool}\\
\mathbf{b}_{2x + i, 2y + i, l} &= \mathrm{ReLu}\bigg(\sum_{i, j\in \{0,1\}}\sum_{k}^{}\mathbf{w}_{i, j, k, l}\cdot \mathbf{a}_{x, y, k} + \mathbf{c}_{l}\bigg),\label{up_sampling}
\end{align} 
where $\mathbf{c}$ is the bias vector, $\mathbf{w}$ the weights matrix, $\mathbf{a}$ the input feature map, $\mathbf{b}$ the output feature map. Here, $x$ is the width, $y$ is the height, $k$ are the features index and $l$ is the depth of image.

\subsection{Optimization via meta-learning}
Since the dataset we used in this work  is small, this justifies that we adopt the transfer learning and a meta-learning  type of optimization. The idea of finding the optimized deep models $\Psi(\theta, \cdot)$ that could be efficiently tested on a small dataset using meta-learning is as follows. From the basic  optimization problem, we want to minimize a loss function  $\mathcal{L}(y, \Psi(\theta, X))$ which uses the cross-entropy for classification problems:
\begin{equation}
    \mathcal{L}(y, \Psi(\theta, X))=-\sum_{i}y_i\log \Psi(\theta, X_i),\label{eq:loss}
\end{equation}
where the vector $y$ with components $y_i$ are know expected predictions, and the vector $X$ with entries $X_i$ are features network inputs. Assuming a sampled task, $\mathcal{T}\sim  \rho(\mathcal{T})$ from the distribution of tasks, $p(\mathcal{T})$. Split the entire dataset into training, validation, and testing sub-datasets and randomly choose some $N$ samples of labels from each of the classes and sequentially update the model weights vectors $\theta^{j}_i$:
\begin{equation}
    \theta^{j}_i=\theta^{(j-1)}_i-\alpha\nabla_{\theta_i}\mathcal{L}(y_{\mathcal{S}_i}, \Psi(\theta, X_{\mathcal{S}_i})),
\end{equation}
where $\alpha$ is the learning rate, $y_{\mathcal{S}_i}$ and $X_{\mathcal{S}_i}$ are the expected input predictions sub-vector and input features sub samples elements of the support set $\mathcal{S}$. Note that $\theta^{0}_i\equiv \theta$. After the  update of task-specific weight $\theta^{j}_i$, the update of $\theta$ now properly follows
\begin{equation}
     \theta \gets \theta-\beta \nabla_{\theta}\sum_{\mathcal{Q}_i\in \mathcal{T}_i}\mathcal{L}(y_{\mathcal{Q}_i}, \Psi(\theta_i^P, X_{\mathcal{Q}_i})).
\end{equation}
Here, $\beta$ is the learning rate, $y_{\mathcal{Q}_i}$ and $X_{\mathcal{Q}_i}$ are the expected input prediction and feature vector elements of the query set $\mathcal{Q}$. It can be seen that the update of $\theta$ is an average of the objective function defined in Equation~\ref{eq:loss} across the query set samples. 

\section{Evaluation metrics}
\label{evaluation}
\subsection{Confusion matrix}
    The study solved a binary classification problem by evaluating the optimal solution to predict whether the wildlife fences are either single fences or double fences based on the characteristics of the wildlife fences. The evaluations are done during training based on confusion matrix~\citep{hossin2015review}. For the confusion matrix, the table row shows the actual class, and the column represents the predicted class. The confusion matrix shows the correspondence between predicted and true values. The cells and columns of the confusion matrix represent the correct and incorrect predictions with summarized count values for each class for all the possible correlations. The cell in the $i^{th}$ row and $j^{th}$ column means the percentage of the $i^{th}$ class samples which are classified into the $j^{th}$ class. The diagonal cell of the matrix contains the number of correctly identified pixels for each class. From the confusion matrix in Table~\ref{tbl:cm}, $tp$ is when the actual class is positive and the predicted class positive. An $fp$ is when the actual class is negative and the predicted class positive. A $tn$ is when the actual class is negative and the predicted class is negative. An $fn$ is when the actual class is positive and the predicted class negative. Both $tp$ and $tn$ represent the number of positive and negative instances that are correctly classified. While $fp$ and $fn$ represent the number of misclassified negative and positive instances.

\subsection{Intersection over Union (IoU)}

IoU, or the Jaccard index, is a metric that determines the location of an object in the image by computing the ground truth $g$  and predicted mask $p$ to determine the IoU. The ground truth is manually annotated on the image, whereas the predicted mask is usually the output of a neural network. The IoU is given as

\begin{align}  
IoU = \frac{\lvert g \cup p \lvert}{\lvert g\cap p \lvert} = \frac{tp}{tp + fp + fn},
\label{iou}
\end{align} 

where $tp$ stands for true positive, $fp$ for false positive, $fn$ for false negative. Further, we compute the Mean Intersection-Over-Union (MIoU), a commonly used evaluation metric in determining the segmentation performance of an image~\cite{li2021application}. The metric is computed as follows:
\begin{align}
MIoU &=\frac{1}{c}\sum_{j=1}^c\frac{ tp_j}{tp_j + fp_j + fn_j} ,
\label{miou}
\end{align}

where $c$ is the total number of classes, IoU always ranges between 0 and 1. The ideal performance is 1, which means that the prediction is the same as ground truth values and vice versa for 0. When the intersection area is larger, IoU will be close to 1.

\subsection{Dice similarity coefficient}
The dice similarity coefficient metric is applied to a variety of datasets to measure the similarity of two sets of data~\cite{zou2004statistical} and is used broadly in image segmentation tasks. The metric is computed as:
\begin{align}  
DSC (g,p) = \frac{2 \times \lvert g \cap p \lvert}{\lvert g \lvert + \lvert p \lvert} = \frac{2tp}{2tp + fp + fn},
\label{dsc}
\end{align}

where $ \lvert g \lvert $ and $ \lvert p \lvert $ are the number of elements in  $g$ and $p$, respectively. and the symbol $\cap$ represents the intersection of two masks.

\subsection{Accuracy, Recall, Precision, F1-Score and Support}
Accuracy is the ratio of correct predictions out of the total of all observations. This measure is the most intuitive. However, it is best when the datasets are symmetric (i.e., values of $fp$ and $fn$ are almost the same):
\begin{align}  
\mathrm{Accuracy} = \frac{tp+fn}{tp+fp+fn+tn},
\label{acc}
\end{align} 

The recall is a measure of the classifier's completeness, the ability of a classifier to find all positive instances correctly. Each class is defined as the ratio of true positives to the sum of true positives and false negatives: 
	\begin{align}
            \mathrm{Recall}=\frac{tp}{fn+fn}.
        \end{align}     
Precision is a measure of a classifier’s exactness. For each class, it is defined as the ratio of true positives to the sum of true and false positives: 
\begin{align}
\mathrm{Precision}=\frac{tp}{fn+tp}.
\end{align}
The F1-score is a weighted harmonic mean of precision and recall such that the best score is 1.0 and the worst is 0.0. Generally speaking, F1-scores are lower than accuracy measures as they embed precision and recall into their computation. As a rule of thumb, the weighted average of F1-score should be used to compare classifier models, not global accuracy. F1-score is usually more useful than accuracy when one have an uneven class distribution. Accuracy works best if false positives and false negatives have similar costs. If the cost of false positives and false negatives are very different, it is better to look at both precision and recall:
\begin{align}
\mathrm{F1}\text{-}\mathrm{score}=\frac{2(r  p)}{(r + p)}.
\end{align}
Support is the number of actual occurrences of the class in the specified dataset. Imbalanced support in the training data may indicate structural weaknesses in the reported scores of the classifier and could indicate the need for stratified sampling or rebalancing. Support does not change between models but instead diagnoses the evaluation process.
    \begin{table}
    \caption{Confusion matrix used as a performance measurement for machine learning classification} \label{tbl:cm}
    \begin{center}
		\begin{tabular}{ |c|c c c| } 
			\hline
			& Predicted Class&  & \\

			\hline
			\multirow{3}{4em}{Actual Class} &  & Yes  &  No \\ 
			\cline{2-4}
			& Yes & True positive ($tp$) & False negative ($fn$) \\ 
			\cline{2-4}
			& No  & False positive ($fp$) & True negative ($tn$) \\ 
			\hline
    \end{tabular}
	\end{center}
\end{table}

\section{Experiments and discussion}
\label{results}
All experiments  in this paper are performed using a computer equipped with an Intel(R) Xeon(R) CPU E5-2695 v2 @ 2.40GHz computing unit, INTEL graphics INTEL 7 CPU for classification, segmentation, and detection tasks. We used the Keras framework with a Tensorflow backend. Results using state-of-the-art networks are compared for all datasets. 

\subsection{Recognition of wildlife fences using aerial and still datasets}
Table~\ref{tbl:disk_size_classify} shows the number of parameters for the different classification architectures used in this paper. We observe that models with smaller parameters outperformed models with a bigger number of parameters. We also argue that models with smaller hard disk space are easily deployed on mobile clients compared to bigger models, which are usually slow. As for Table~\ref{tbl:classify_acc_loss_results}, the CNN model achieved a classification test accuracy of 73\% in aerial images and 82\% in still images. However, the ResNet model showed a better performance on test set of 100\% on aerial images and 95\% on still images. This is justified by the transferable capabilities of ResNet from one adaptation domain to another, and their robustness to operate on small data sets.
Results indicate the CNN model performed poorly compared to ResNet. However, results also show that CNN performed better on aerial images than ResNet on still images. Therefore, we recommend using transfer learning on the N2 wildlife fences dataset compared to CNN's because transfer learning architecture performs better than CNN. In addition, the test accuracy using CNN is generally higher than train and validation accuracy. We attribute this to the small dataset used in this study and the inability of the CNN model to be robust compared to ResNet. Due to these weaknesses, we computed more metrics such as confusion matrix, precision, recall, F1-score, and support, with their associated average accuracy, macro average, and weighted average.

\begin{figure*}
\begin{multicols}{2}
   \textbf{(a) Aerial, CNN} \par \includegraphics[width=\linewidth]{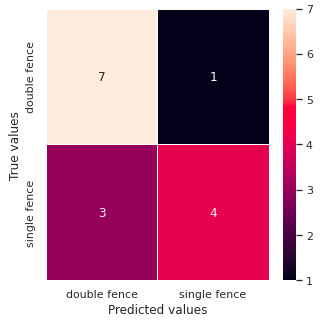}\par 
    \textbf{(b) Aerial, ResNet} \par\includegraphics[width=\linewidth]{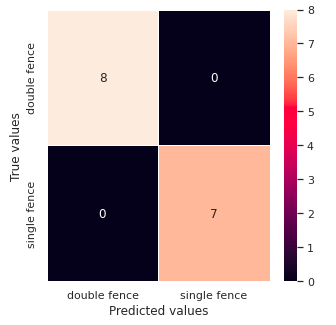}\par 
    \end{multicols}
\begin{multicols}{2}
    \textbf{(c) Still, CNN} \par\includegraphics[width=\linewidth]{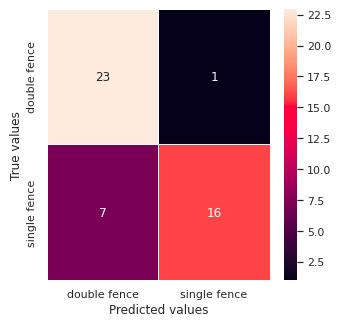}\par
     \textbf{(d) Still, ResNet} \par \includegraphics[width=\linewidth]{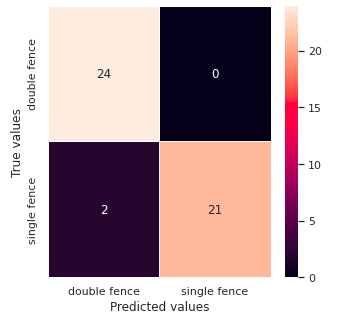}\par
\end{multicols}
\caption{Four confusion matrix showing (a) images from drone classified using CNN, (b) images from drone classified using ResNet, (c) images from standalone camera classified using CNN, and (d) images from standalone camera classified using ResNet}
\label{cm(s)}
\end{figure*}

For the confusion matrix in Figure~\ref{cm(s)} (a), (b), (c), and (d) correspond to the results from aerial images with CNN, aerial images with ResNet, still images with CNN, and still images with ResNet respectively. Figure~\ref{cm(s)} (a) shows 7 double fences classified as double fences and 1 double fence misclassified as single fences, as well as 4 single fences classified as single fences and 3 single fences classified as double fences. For Figure~\ref{cm(s)} (b), we observe classifications of 8 double fences and 7 single fences with no misclassifications. Results of the confusion matrix when using still images with CNN and ResNet achieved acceptable results with 23 double fences and 16 single images predicted correctly as double fences and single fences, respectively. In the same Figure~\ref{cm(s)} (c), 7 single fences are predicted wrongly as double fences, and 1 double fence is incorrectly classified as a single fence. Finally, Figure~\ref{cm(s)} (c) results are from using still images and the ResNet model. The model achieved correct predictions of 24 double fences and 21 single fences. The model also misclassified 2 single fences as double fences, with no misclassifications for double fences. 
\begin{table}
\caption{Data reconstructed for segmentation and detection tasks}
\begin{center}
\begin{tabular}{llll}
\multicolumn{1}{c}{\bf Dataset name}&\multicolumn{1}{c}{\bf Model} &\multicolumn{1}{c}{\bf Disk size (MBs)} &\multicolumn{1}{c}{\bf Parameters}
\\ 
\hline
\hline
Aerial &CNN & 354.0 & 29,492,922\\
Aerial & ResNet & 101.3& 24,125,208\\
Still & CNN &393.3 & 32,769,882\\
Still & ResNet &101.3 &24,125,208\\

\hline
\label{tbl:disk_size_classify}
\end{tabular}
\end{center}
\end{table}

As in Table~\ref{tbl:cf(s)}, the CNN model achieved the classification average accuracy of 73\% on aerial images and 83\% on still images. However, the ResNet model performance is of 100\% on aerial images as well as 96\% on still images. 
Generally, the classification models performed better on aerial images than still images because the drone can capture wildlife fences from the front view, back view, and top view, while the standalone camera captures only by front view. Meaning drones capture more details of the fence and its associated features as compared to a standalone camera. Based on these results, we recommend the use of transfer learning on the wildlife fences dataset as compared to CNN's when using classification algorithms. We further attribute the poor performance to the smaller datasets used for both aerial and still.

\begin{table}
\caption{Results from the classification task for both aerial (dataset 1) dataset as well as still (dataset 2) dataset}
\begin{center}
\begin{tabular}{llllllll}
\multicolumn{1}{c}{\bf \makecell{Dataset\\ name}}&\multicolumn{1}{c}{\bf Model} &\multicolumn{1}{c}{\bf  \makecell{Train \\ acc \\ (80\%)}} &\multicolumn{1}{c}{\bf \makecell{Train \\ loss}} &\multicolumn{1}{c}{\bf  \makecell{ Val \\ acc \\ (10\%)}} &\multicolumn{1}{c}{\bf  \makecell{Val \\ loss}} &\multicolumn{1}{c}{\bf  \makecell{Test \\ acc  \\ (10\%)}} &\multicolumn{1}{c}{\bf  \makecell{Test \\ loss}}
\\ 
\hline
\hline
\makecell{Aerial} & CNN &74.81& 0.6654& 73.33& 6579&73.33& 0.7006\\
\hline
\makecell{Aerial}& ResNet &100& 4.6209e-06&100& 0.0054& 100& 3.5681e-04\\
\hline
\makecell{Still} &CNN &72.37&0.5996&61.70&1.1490&82.98&0.3767\\
\hline
\makecell{Still} &ResNet&100&1.6472e-08&97.87&0.0275&95.74&0.3922\\
\hline
\label{tbl:classify_acc_loss_results}
\end{tabular}
\end{center}
\end{table}

\subsection{Prediction and detection of wildlife fence insulators}
\begin{figure}
		\centering
		\includegraphics[width=\textwidth]{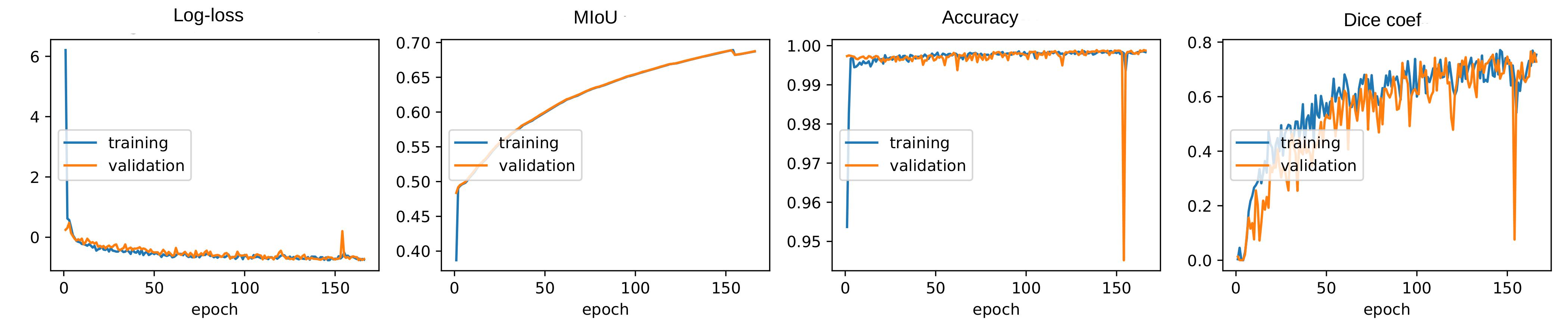}
		\caption{ Loss, MIoU, accuracy and dice curves of the drone image dataset model}\label{fig:plot_losses_drone}
	\end{figure}

\begin{figure}
		\centering
		\includegraphics[width=\textwidth]{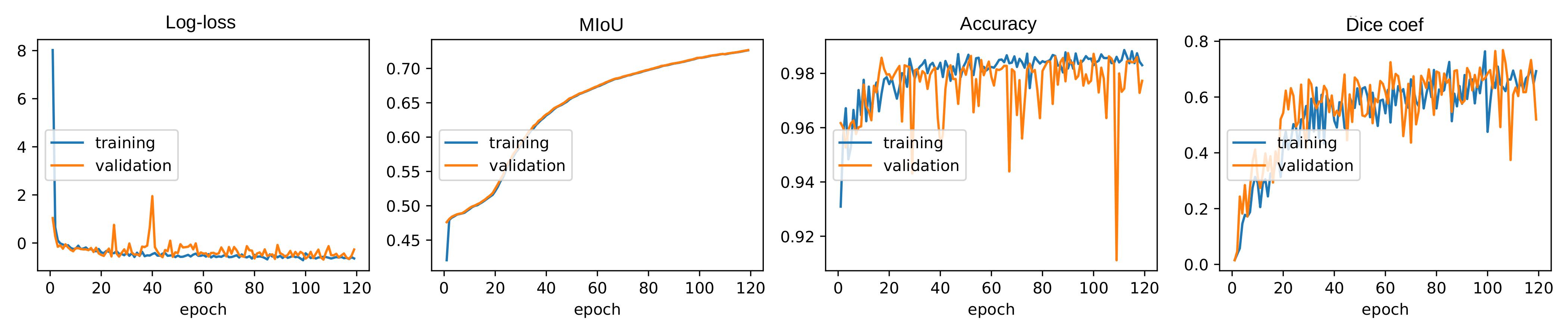}
		\caption{ Loss, MIoU, accuracy and dice curves of the still image dataset model}\label{fig:plot_losses_still}
	\end{figure}

On average, 12 days and 2 hours are spent on training the aerial and still datasets. During training, the loss, MIoU, accuracy, and dice coefficient are calculated on the training and validation datasets at the end of each epoch. And the parameters that got in the final epoch while training is considered the training results. All the parameters of the training process get stored in a dictionary, and those values are retrieved and used for plotting the graphs. These graphs are used to evaluate the performance of the model. The model is said to be good if the epoch vs accuracy graph is exponential and if the epoch vs loss graph is a hyperbolic curve. Figures~\ref{fig:plot_losses_drone} and 
 \ref{fig:plot_losses_still} show the loss, MIoU, accuracy, and dice coefficient curves of the drone and still datasets, respectively. The model results are in Table~\ref{tbl:results}. In both datasets, the loss curves show some slight oscillations when computing the loss, accuracy, and dice coefficient metrics. This is related to the small mini-batch size set in this study. The network cannot find the convergence direction well at the early stage of training. When we observe the loss curves, the network gradually becomes stable with continuous network training obtaining results in Table~\ref{tbl:results}. 

{\scriptsize 
\begin{longtable}{ p{3cm} |  p{2cm} | p{1cm} | p{1.5cm} | p{1.5cm} |p{1.3cm} |p{1.3cm}|p{1.3cm}} \caption{Aerial and Still datasets fine-tuned on CNN and ResNet binary classification algorithms. The wildlife single and double fences classes evaluations are precision, recall, F1-score, support, average accuracy, macro average, and weighted average.}
\label{tbl:cf(s)}\\
\hline
\multicolumn{1}{c}{\bf \makecell{Avg\\ metrics}}&\multicolumn{1}{c}{\bf \makecell{Dataset}}&\multicolumn{1}{c}{\bf \makecell{Model}}&\multicolumn{1}{c}{\bf \makecell{Fence \\ type}}&\multicolumn{1}{c}{\bf Precision} &\multicolumn{1}{c}{\bf \makecell{Recall}} &\multicolumn{1}{c}{\bf \makecell{F1-score}} &\multicolumn{1}{c}{\bf \makecell{Support}}
\\ 
\hline
\hline
 & Aerial & CNN  & double fence & 0.70 & 0.88 & 0.78&  8\\
 & & &             single fence & 0.80 & 0.57 & 0.67 & 7 \\
 Accuracy &&&                        &       &       & 0.73 & 15 \\
 Macro Avg &&&                  &  0.75 & 0.72   & 0.72 & 15 \\                 
Weighted Avg &&&                  &  0.75 & 0.73   & 0.73 & 15 \\               \hline

 & Aerial & ResNet  & double fence & 1.00 & 1.00 & 1.00 &  8\\
 & & &                single fence & 1.00 & 1.00 & 1.00 & 7 \\
 Accuracy &&&                        &       &       & 1.00 & 15 \\
 Macro Avg &&&                  &  1.00 & 1.00  & 1.00 & 15 \\                 
Weighted Avg &&&                  &  1.00 & 1.00  & 1.00 & 15 \\                

\hline
 & Still & CNN  & double fence & 0.77 & 0.96 & 0.85 &  24\\
 & & &            single fence & 0.94 & 0.70 & 0.80 & 23 \\
 Accuracy &&&                        &       &       & 0.83 & 47 \\
 Macro Avg &&&                  &  0.85 & 0.83 & 0.83 & 47 \\                 
Weighted Avg &&&                  & 0.85 & 0.83 & 0.83 & 47 \\                

\hline
 & Still & ResNet  & double fence & 0.92 & 1.00 & 0.96 &  24\\
 & & &               single fence & 1.00 & 0.91 & 0.95 & 23 \\
 Accuracy &&&                        &       &       & 0.96 & 47 \\
 Macro Avg &&&                  &  0.96 & 0.96 & 0.96 & 47 \\                 
Weighted Avg &&&                  & 0.96 & 0.96 & 0.96 & 47 \\                

\hline
\end{longtable}}

\subsection{Mask prediction accuracy}
We applied the U-Net model to the wildlife aerial image dataset and wildlife still dataset, which is labeled as insulators and not insulators. Every image of the data is RGB and has 512 $\times$ 512 pixels resolution. 
The best MIoU are recorded for training, validation, and testing for both aerial and still datasets in Table~\ref{tbl:results}. Adam with learning is adopted as an optimization algorithm~\cite{kingma2014adam}.

\begin{figure}
		\centering
		\includegraphics[width=\textwidth]{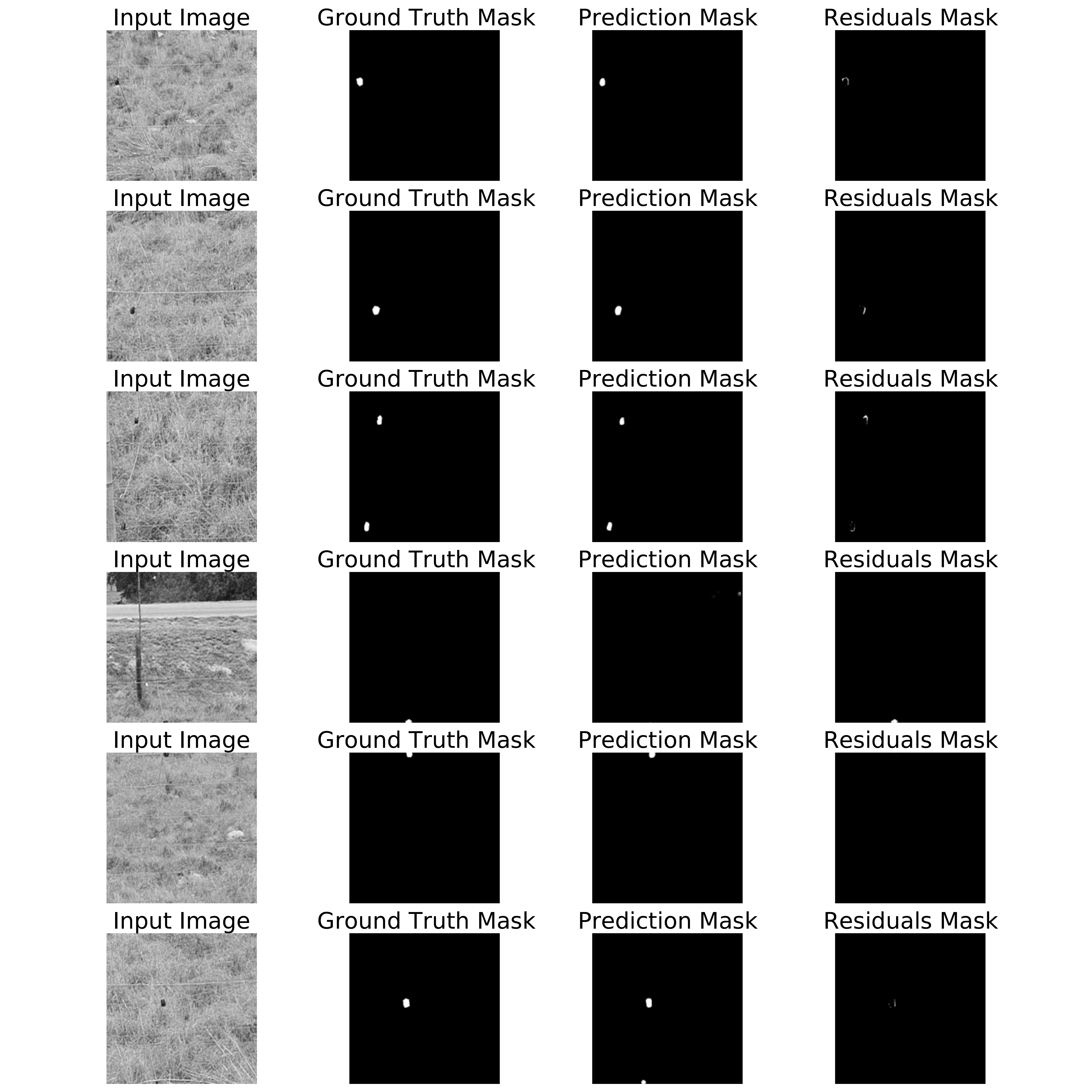}
		\caption{Representation of actual images, ground truth masks, predicted masks, and residual masks on aerial dataset}		\label{fig:masks}
	\end{figure}

The output of the given neural network contains an
image where each pixel corresponds to a probability to detect
interested area. The size of the output image coincides with the input image. To have only binary pixel values, we choose a threshold of 0.5. For all pixel values below the specified threshold, we set it to zero, while for all values above the threshold, we set it to 1. We then multiplied the results of every pixel in an output image by 255 and got a black-and-white predicted mask. Further, for each predicted ground truth mask, we calculate the residuals. The smaller the residuals, the better the model performance, and the more the residuals, the poor the model performance, as shown in Figure \ref{fig:masks}

{\scriptsize 
\begin{longtable}{ p{3cm} |  p{3cm} | p{1cm} | p{1.5cm} | p{1.5cm} |p{1.5cm} |p{1.5cm}} 
\caption{Prediction results on training, validation and testing dataset samples. Still, dataset performed better on small batches than on bigger batches. The aerial slightly performed better on the 16 batch size than the 8 batch size. Generally, the model performed better on the drone dataset than still images.}
\label{tbl:results}\\
\hline
\multicolumn{1}{c}{\bf Dataset}&\multicolumn{1}{c}{\bf \makecell{Dataset \\ sample}}&\multicolumn{1}{c}{\bf \makecell{Batch \\ size}} &\multicolumn{1}{c}{\bf Loss} &\multicolumn{1}{c}{\bf \makecell{Mean IoU}} &\multicolumn{1}{c}{\bf \makecell{Accuracy}} &\multicolumn{1}{c}{\bf \makecell{Dice coef}}
\\ 
\hline
\hline
Drone & Train & 16  & -0.6024 & 0.6776 &0.9980 & 0.6361  \\
\cline{3-7}
& & 8 & -0.5492 &  0.6442& 0.9974 & 0.5768  \\
\cline{2-7}
 & Val & 16  & -0.7097& 0.7322 &0.9987 & 0.7066  \\
 \cline{3-7}
& & 8 & -0.5595 &  0.6989& 0.9977 & 0.5793  \\
\cline{2-7}
 & Test & 16  & -0.6279& 0.7345 &0.9982 & 0.6771 \\
 \cline{3-7}
& & 8 & -0.5938 &  0.6992& 0.9981 & 0.6378  \\
\hline
Still & Train & 16  & 22.7265 & 0.4602&0.9684 & 0.0053  \\
\cline{3-7}
& & 8 & -0.1184 & 0.5017& 0.9646 & 0.1872  \\
\cline{2-7}
 & Val & 16  & 22.7265& 0.4857 &0.9698 & 0.0097  \\
 \cline{3-7}
& & 8 & -0.0826 &  0.5405& 0.9678 & 0.1644  \\
\cline{2-7}
 & Test & 16  & 19.7826& 0.4856 &0.9740 & 0.0047 \\
 \cline{3-7}
& & 8 & -0.0580&  0.5388& 0.9704 & 0.1069 \\
\hline
\end{longtable}}

\subsection{Insulator detection with OpenCV}
\begin{figure}
		\centering
		\includegraphics[width=15cm, height=15cm]{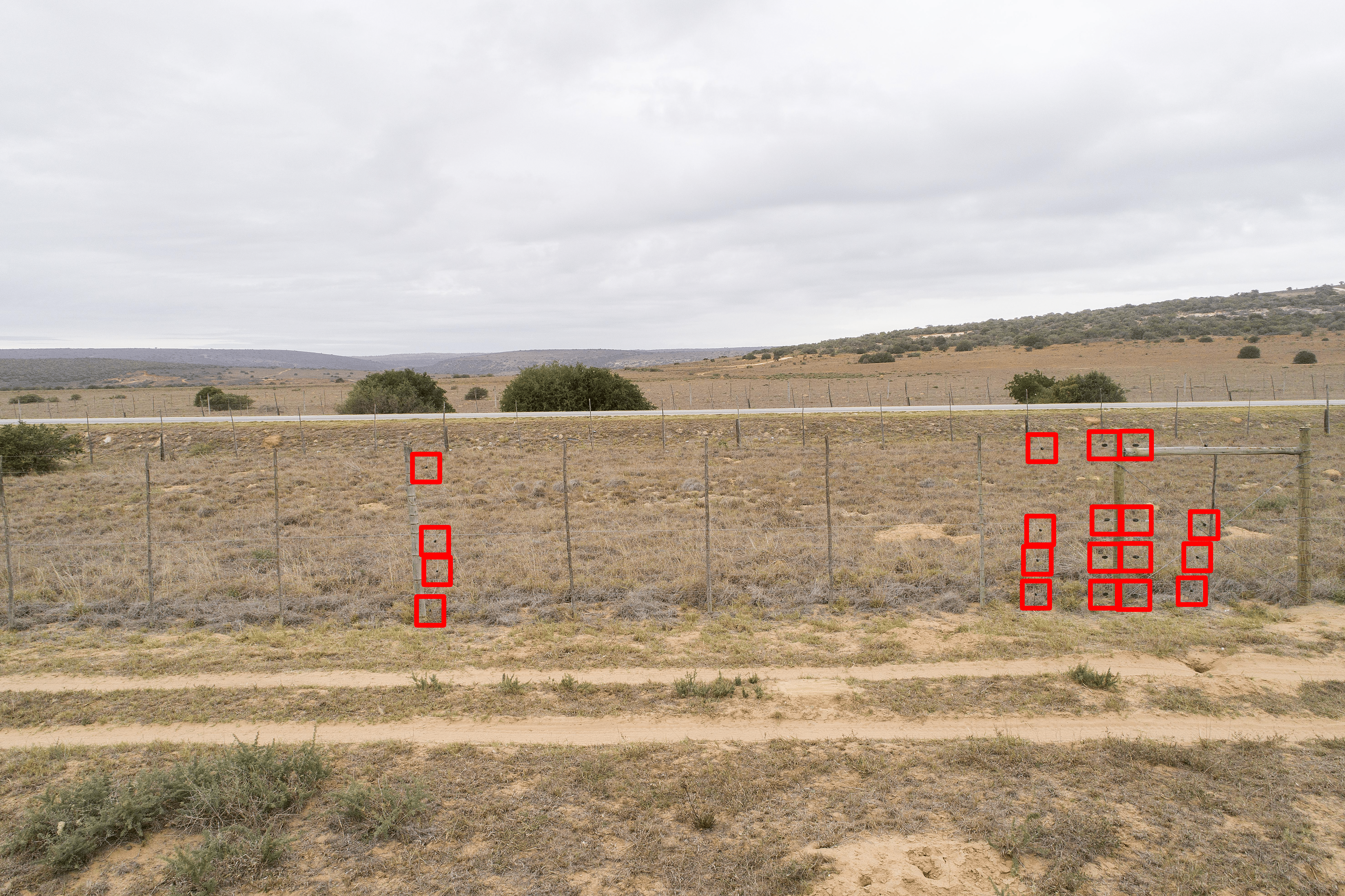}
		\caption{Detected insulators on wildlife electric fence at Amakhala private game reserve from the drone dataset}		\label{fig:detected_predicted_masks_aerial}
	\end{figure}

\begin{figure}
		\centering
		\includegraphics[width=15cm, height =15cm]{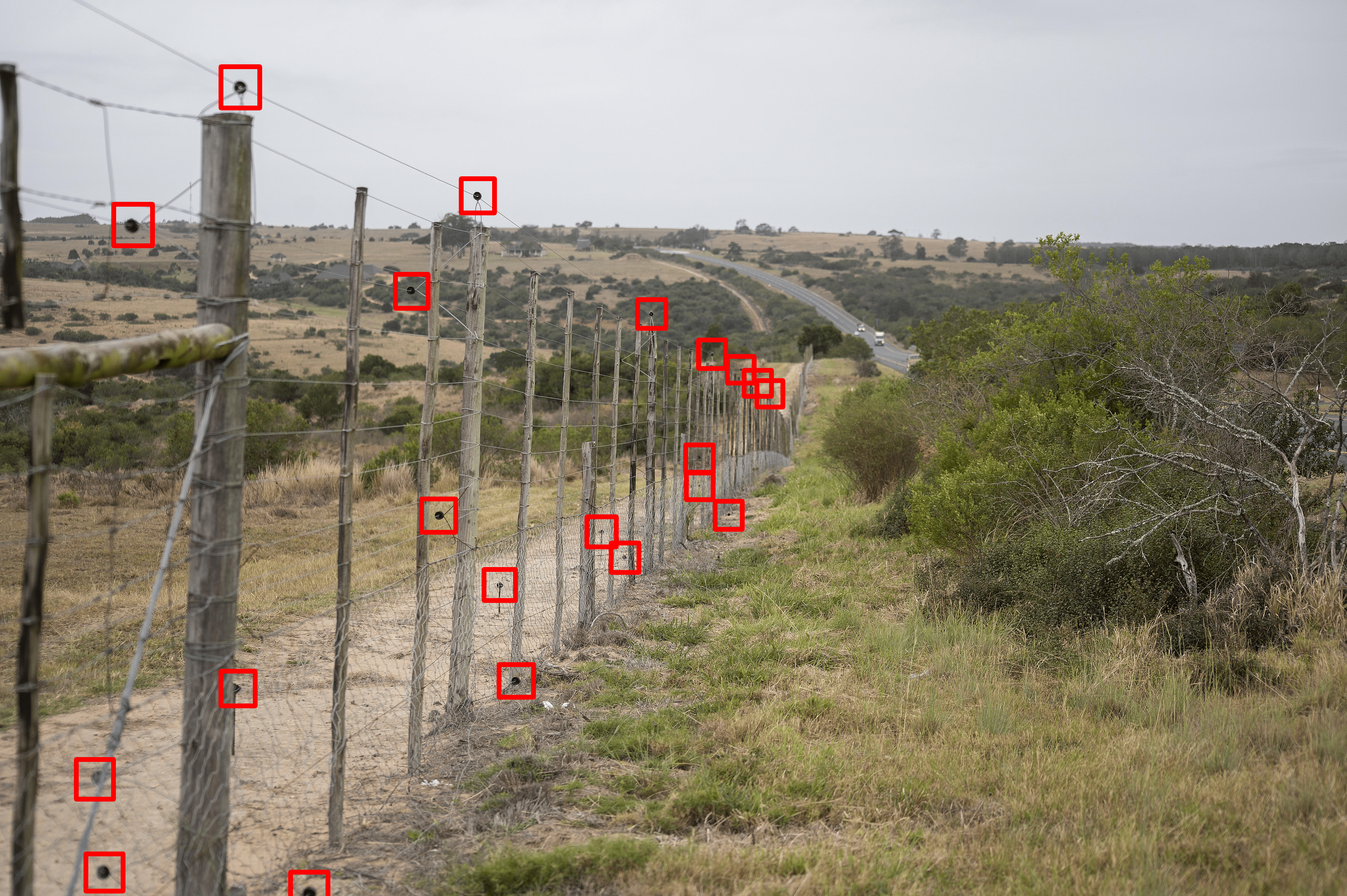}
		\caption{Detected insulators on wildlife electric fence at Amakhala private game reserve from the still dataset}		\label{fig:detected_predicted_masks_still}
	\end{figure}

OpenCV is an open-source library widely used for image analysis, recognition, and deep learning. The software is used exhaustively to recognize objects in photos and videos. Object detection is a subset of signal processing, big data, and machine learning focusing on identifying features in pictures and videos. Images are labeled as positive and negative to train a cascade function, which is used to detect objects in images~\cite{viola2001rapid}. In this work, we detect insulators using OpenCV as shown in Figure~\ref{fig:detected_predicted_masks_aerial} from the aerial dataset by finding contours defined by joining lines of all points at the image boundary with the same intensity. OpenCV uses a function findContours() to extract contours from an image that meets a specified threshold. In addition, a boundingRect() function is adopted to approximate rectangles of the blobs on an image to show the area of interest after getting the image’s contours or outer shape. This function returns four values, the x-coordinate, y-coordinate, width, and height of the blob. These values are used to draw a rectangle on the image part using pixel coordinates. A few pixels are added to these values to draw a slightly bigger box. These coordinates from mask images are inferred into the actual images to draw the borders around them. The cut-outs images and their corresponding mask images are passed through our OpenCV code to detect all the sources in the image. Later these are concatenated back to an original image of which borders are sketched around it, as shown in Figures~\ref{fig:detected_predicted_masks_aerial} and \ref{fig:detected_predicted_masks_still}, applied to the drone and still images respectively.

\newpage
\subsection{Discussion}
\label{discussion}
Wildlife-vehicle collision incidences are high in South Africa in areas with wildlife fences~\cite{eloff2008temporal}. This prompts research on how to minimize and prevent wildlife collisions with cars and ensure the safety of drivers, pedestrians, and other road users. Ecologists and wildlife conservationists in Africa want an easy way to detect WVC areas with high accuracy and low cost while minimizing the weaknesses of roadkill prediction using time series analysis, an approach currently adopted in road ecology in detecting roadkill areas. To answer the ecologists, this study automatically investigates the possibility of the U-Net segmentation algorithm, CNN, and ResNet in detecting and recognizing WVC areas. It does this by understanding the performance of CNN, ResNet, and U-Net trained on still and aerial wildlife fence and wildlife fence insulator datasets. Ultimately, for segmentation, it verifies if this is a good approach for detecting WVC fences by further drawing bounding boxes on the predicted insulators as shown in Figure~\ref{fig:detected_predicted_masks_aerial}. In addition, the ResNet and CNN recognize WVC areas using wildlife single and double fences shown on the confusion matrix in Figure~\ref{cm(s)} as well as precision, recall, F1-score, and support in Table~\ref{tbl:cf(s)}. Generally, deep learning models' accuracy is correlated to the size of the dataset; the bigger the dataset, the higher the accuracy and vice-versa~\cite{tabak2019machine, willi2019identifying}. This is un-realist in an ecological sense since large amounts of training data can not be easily produced by ecological researchers. For that purpose, we designed the U-Net model for smaller ecological wildlife fences insulator datasets. And adopted ResNet for the classification task, which performs better than CNN.

Our model results in Table~\ref{tbl:results} show that the U-Net model performs better on the aerial dataset than the still dataset, even though the aerial dataset is small. This is partially true due to the variations in specific camera parameters, which are known to affect the performance and generalization of deep learning models~\cite{liuZhenyi2020}. The two camera devices, DJI Phantom 4 Pro UAV drone and Nikon Z6 Mk 2 still camera, have different parameters affecting the U-Net, CNN, and ResNet results. Also, the drone captures more details (front, back, and top views) for machine learning models to learn compared to the still camera, which captures only the front view. In addition, the images are over-zoomed or decompressed, leading to poor performance since we have much information lost.

Our work can uniquely combine ecological wildlife fences, wildlife fence insulator data, and deep learning techniques. The field of image recognition, especially when applied to wildlife fencing in detecting WVC areas, is still young. While most researchers have used similar methodologies, they have used them in different fields but not in detecting WVC areas using wildlife fences to minimize WVC.

Our results demonstrate the successful capabilities of ResNet, CNN, and U-Net within the ecological domain, although with identified limitations. The algorithms, if deployed, can be a powerful and practical tool that helps ecologists extract ecological information from the wildlife fencing algorithm deployed. 


\subsection{Future Scope}
We offer a few possibilities for future work by combining traditional wildlife fencing, classification, and segmentation algorithms.

First, we acknowledge that our datasets are not distributed. It comprises only two game reserves in Grahamstown, Eastern Cape, South Africa. As a result, future work should repeat the data collection process by representing more WVC areas in different provinces of South Africa and testing the model's generalization between the different identified locations. This will improve machine learning systems in generalizing their training data, which makes the model better in generalizing~\cite{lecun2015deep}.

Second, deploying machine learning models to the client devices is vital, as periodically retraining the model from model staleness. Deployment has been successful in public policy~\cite{ackermann2018deploying}, health~\cite{kulkarni2021key}, education~\cite{takabe2013rapid}, and banking~\cite{tabiaa2019deployment}

Thirdly, to further improve the classification accuracy of remote sensing, adopting image localization and detection algorithms might help to extract the location of an object in an image using algorithms such as histogram of oriented gradients, YOLO, Region-based Convolutional Neural Networks (R-CNN), Fast R-CNN, and Faster R-CNN.

Fourthly, Building more images of the dataset using synthetic images will improve performance by exponentially increasing the number of example images a model sees during training. This approach is widely used by U-Net with generative adversarial networks and is successful~\cite{schonfeld2020u, wu2019u}

Fifthly, models developed using aerial and still datasets may be continuously trained and deployed to minimize model staleness by using proactive training~\cite{prapas2021continuous}. This approach will improve model performance and reliability by retraining the images the model provided as incorrect predictions. 

Lastly, we believe the future of deep learning models relies upon developing an end-to-end WVC fence detection system to monitor and detect WVC areas. The system should detect wildlife fences as WVC areas and alert the driver. Further, the system should allow the driver to save images and GPS. The collected image dataset can be used to retrain the model. And the GPS will help improve the existing roadkill areas detection system by detecting future patterns and distribution of WVC areas using time series analysis.

\section{Conclusion}
\label{conclusion}
The study aimed to evaluate the model performance of modern computer vision algorithms applied to aerial and still image datasets. For this purpose, we utilized drones and standalone cameras to obtain images of wildlife fences on N2 highways. These images were sorted as single and double fences, each associated with the device from which they originate, i.e., drone or standalone camera. For the classification task, the images were labeled and fed as input to the computer vision algorithms to recognize fences. In addition to the dataset contribution, we also conducted a comparative study on state-of-the-art image-based classification algorithms and reported our experimental results in this paper. We exploited CNN and ResNet50 architecture for the recognition tasks. The recognition task detects the test image as a single or double fence.

For the segmentation task, our contribution is the dataset that contains segmented images and masks for wildlife electric fence insulators. Our proposed approach to detecting fences facilitates the development of image segmentation algorithms with wildlife fences and their associated features. This is done by annotating the dataset, which increases the efficiency of machine learning projects in detecting areas of interest. Detecting wildlife fence features such as insulators is essential in mitigating WVC.

Comparisons of CNN results and those from the ResNet50 architecture showed that ResNet50 outperforms CNN when applied to aerial and still images. Generally, the classification and U-Net models performed better on aerial images than still images because the drone can capture wildlife fences by the front view, back view, and top view, while the standalone camera captures only by front view. Meaning drones capture more details of the fence and its associated features as compared to a standalone camera. Besides, image features from the still camera are over-decompressed compared to images from a drone camera. That means more information is lost, which has contributed to the weak performance of these models.

In the future, we plan to gradually enlarge our dataset for single and double fences using drone and standalone devices. We hope making public our dataset to foster future image-based wildlife fence recognition research. Further, we will minimize the decompression of images, detect other vital features of wildlife electric fences, adopt localization and detection algorithms to detect the location of objects in an image, and deploy the algorithms on client devices. Besides, we shall also adopt using other segmentation techniques.

\section*{Disclosure statement}
No potential conflict of interest was reported by the author(s)

\section*{Ethical clearance}
Ethical review and approval were waived for this study by the Animal Research Ethics Committee of Rhodes University, South Africa, because the study did not involve the use of live animals and therefore did not require ethical clearance.

\section*{Funding}
This study was carried out with financial support from Rhodes University.

\section*{Data availability}
The data that support the findings of this study are accessible from the corresponding author upon request.

\section*{Acknowledgment}
The authors would like to thank Amkhala and Lalibela game reserves in South Africa for many valuable discussions and educational help in wildlife fence features, on how wildlife fence works, and for allowing us to take photographs of the wildlife fence on highways. The authors also express sincere gratitude to HLB Photography \footnote{\url{https://www.hlbphotography.co.za/}} for providing us with a photographer and drone pilot to capture pictures.

\section*{Authors’ contributions}
Conceptualization, I.N. and M.A.; methodology, I.N. and M.A.; data collection I.N., M.A. and P.O.; data curation, I.N.; software, I.N.; validation, I.N., M.A. and M.N.; formal analysis, I.N., M.A., P.O.,  M.N., J.L.E.K.F. and F.T.; investigation, I.N.; writing—original draft preparation, I.N.; writing—review and editing, I.N., M.A., P.O., M.N., J.L.E.K.F. and F.T.; visualization, I.N., M.A., P.O., J.L.E.K.F. and F.T.; supervision, M.A. and P.O.; editing the document with an ecological audience in mind, I.N. and M.N.; project administration, I.N. and M.A.; funding acquisition, M.A. All authors wrote the manuscript, commented and improved it. All authors read and approved the final version of the manuscript.



\bibliography{elsarticle-template}

\begin{thebibliography}{10}
\expandafter\ifx\csname url\endcsname\relax
  \def\url#1{\texttt{#1}}\fi
\expandafter\ifx\csname urlprefix\endcsname\relax\def\urlprefix{URL }\fi
\expandafter\ifx\csname href\endcsname\relax
  \def\href#1#2{#2} \def\path#1{#1}\fi

\bibitem{PEKOR}
A.~Pekor, J.~R. Miller, M.~V. Flyman, S.~Kasiki, M.~K. Kesch, S.~M. Miller,
  K.~Uiseb, V.~{van der Merve}, P.~A. Lindsey,
  \href{https://www.sciencedirect.com/science/article/pii/S0006320718306578}{Fencing
  africa's protected areas: Costs, benefits, and management issues}, Biological
  Conservation 229 (2019) 67--75.
\newblock \href
  {http://dx.doi.org/https://doi.org/10.1016/j.biocon.2018.10.030}
  {\path{doi:https://doi.org/10.1016/j.biocon.2018.10.030}}.
\newline\urlprefix\url{https://www.sciencedirect.com/science/article/pii/S0006320718306578}

\bibitem{https://doi.org/10.1111/1365-2664.12415}
S.~M. Durant, M.~S. Becker, S.~Creel, S.~Bashir, A.~J. Dickman, R.~C.
  Beudels-Jamar, L.~Lichtenfeld, R.~Hilborn, J.~Wall, G.~Wittemyer,
  L.~Badamjav, S.~Blake, L.~Boitani, C.~Breitenmoser, F.~Broekhuis,
  D.~Christianson, G.~Cozzi, T.~R.~B. Davenport, J.~Deutsch, P.~Devillers,
  L.~Dollar, S.~Dolrenry, I.~Douglas-Hamilton, E.~Dröge, E.~FitzHerbert,
  C.~Foley, L.~Hazzah, J.~G.~C. Hopcraft, D.~Ikanda, A.~Jacobson, D.~Joubert,
  M.~J. Kelly, J.~Milanzi, N.~Mitchell, J.~M'Soka, M.~Msuha, T.~Mweetwa,
  J.~Nyahongo, E.~Rosenblatt, P.~Schuette, C.~Sillero-Zubiri, A.~R.~E.
  Sinclair, M.~R. Stanley~Price, A.~Zimmermann, N.~Pettorelli,
  \href{https://besjournals.onlinelibrary.wiley.com/doi/abs/10.1111/1365-2664.12415}{Developing
  fencing policies for dryland ecosystems}, Journal of Applied Ecology 52~(3)
  (2015) 544--551.
\newblock \href {http://dx.doi.org/https://doi.org/10.1111/1365-2664.12415}
  {\path{doi:https://doi.org/10.1111/1365-2664.12415}}.
\newline\urlprefix\url{https://besjournals.onlinelibrary.wiley.com/doi/abs/10.1111/1365-2664.12415}

\bibitem{hanophy2009fencing}
W.~Hanophy, Fencing with wildlife in mind, XColorado Division of Wildlife,
  2009.

\bibitem{ZAlawcommission}
S.~A.~L. COMMISSION, Uniform national legislation on the fencing of national
  roads,
  \url{https://www.gov.za/sites/default/files/gcis_document/201409/ip16prj1192000.pdf},
  Accessed 15 September 2022 (2000).

\bibitem{WendyCollision2015}
J.~Kioko, C.~Kiffner, N.~Jenkins, W.~J. Collinson, Wildlife roadkill patterns
  on a major highway in northern tanzania, African Zoology 50 (2015) 17--22.

\bibitem{AitorAlmeida2018}
A.~Almeida, G.~Azkune, Predicting human behaviour with recurrent neural
  networks, Applied Sciences 8.
\newblock \href {http://dx.doi.org/https://doi.org/10.3390/app8020305}
  {\path{doi:https://doi.org/10.3390/app8020305}}.

\bibitem{nandutu2022error}
I.~Nandutu, M.~Atemkeng, N.~Mgqatsa, S.~Toadoum~Sari, P.~Okouma,
  R.~Rockefeller, T.~Ansah-Narh, J.~L. Ebongue Kedieng~Fendji, F.~Tchakounte,
  Error correction based deep neural networks for modeling and predicting south
  african wildlife--vehicle collision data, Mathematics 10~(21) (2022) 3988.

\bibitem{defender}
T.~White, Defenders of wildlife. watch out for wildlife facts,
  \url{https://defenders.org/sites/default/files/publications/collision_facts_and_figures.pdf},
  Accessed 5 June 2020 (2020).

\bibitem{federal2018}
C.~Tony, C.~Brian~L, F.~Adam, H.~Marcel, L.~Bruce~F, W.~Bethanie, W.~Chuck,
  Wildlife-vehicle collision reduction study: Report to congress, Tech. rep.,
  U.S Department of Transportation (2008).

\bibitem{Heigl2017}
F.~Heigl, K.~Horvath, G.~e.~a. Laaha, Amphibian and reptile road-kills on
  tertiary roads in relation to landscape structure: using a citizen science
  approach with open-access land cover data, BMC Ecol 17.
\newblock \href {http://dx.doi.org/https://doi.org/10.1186/s12898-017-0134-z}
  {\path{doi:https://doi.org/10.1186/s12898-017-0134-z}}.

\bibitem{WendyCollision2012}
W.~Collision, Roadkill study highlights wildlife road deaths,
  \url{https://www.bridgestone.co.za/news-article/698/roadkill-study-highlights-wildlife-road-deaths},Accessed
  23 April 2020 (2012).

\bibitem{RoadKill}
Wheels24, Roadkill: Why so many animals die on sa’s roads,
  \url{https://www.wheels24.co.za/News/Roadkill-in-SA-Distraction-not-speed-to-blame-20150422},Accessed
  9 April 2020 (2015).

\bibitem{WendyJ2017}
W.~Collinson, H.~Davies-Mostert, W.~Davies-Mostert, Effects of culverts and
  roadside fencing on the rate of roadkill of small terrestrial vertebrates in
  northern limpopo, south africa, Conservation Evidence 14 (2017) 39--43.

\bibitem{s22072478}
I.~Nandutu, M.~Atemkeng, P.~Okouma,
  \href{https://www.mdpi.com/1424-8220/22/7/2478}{Intelligent systems using
  sensors and/or machine learning to mitigate wildlife-vehicle collisions: A
  review, challenges, and new perspectives}, Sensors 22~(7).
\newblock \href {http://dx.doi.org/10.3390/s22072478}
  {\path{doi:10.3390/s22072478}}.
\newline\urlprefix\url{https://www.mdpi.com/1424-8220/22/7/2478}

\bibitem{MarcelP.Huijser2006}
M.~P. Huijser, P.~T. McGowen, W.~Camel, et~al., Animal vehicle crash mitigation
  using advanced technology phase i: review, design, and implementation., Tech.
  rep., Montana State University (2006).

\bibitem{mark2003}
M.~Desholm, Thermal animal detection system (tads). development of a method for
  estimating collision frequency of migrating birds at offshore wind turbines,
  Tech. rep., National Environmental Research Institute (2003).

\bibitem{marcel2012}
M.~P. Huijser, C.~Haas, K.~R. Crooks, et~al., The reliability and effectiveness
  of an electromagnetic animal detection and driver warning system., Tech.
  rep., Colorado. Dept. of Transportation. Research Branch (2012).

\bibitem{CATLETT201962}
C.~Catlett, E.~Cesario, D.~Talia, A.~Vinci,
  \href{https://www.sciencedirect.com/science/article/pii/S157411921830542X}{Spatio-temporal
  crime predictions in smart cities: A data-driven approach and experiments},
  Pervasive and Mobile Computing 53 (2019) 62--74.
\newblock \href {http://dx.doi.org/https://doi.org/10.1016/j.pmcj.2019.01.003}
  {\path{doi:https://doi.org/10.1016/j.pmcj.2019.01.003}}.
\newline\urlprefix\url{https://www.sciencedirect.com/science/article/pii/S157411921830542X}

\bibitem{Butt2021}
U.~Butt, S.~Letchmunan, F.~Hassan, M.~Ali, A.~Baqir, T.~W. Koh, H.~Sherazi,
  Spatio-temporal crime predictions by leveraging artificial intelligence for
  citizens security in smart cities\href
  {http://dx.doi.org/10.20944/preprints202102.0172.v1}
  {\path{doi:10.20944/preprints202102.0172.v1}}.

\bibitem{hp2016roadkill}
P.~HP-Ruffino, K.~G. Rodr{\'\i}guez-C, L.~A. Ramazzotto, P.~D. Freitas, et~al.,
  Roadkill hotspots in a protected area of cerrado in brazil: planning actions
  to conservation, Revista MVZ C{\'o}rdoba 21~(2) (2016) 5441--5448.

\bibitem{ozcan2017identifying}
A.~U. {\"O}ZCAN, N.~K. {\"O}ZKAZAN{\c{C}}, Identifying the hotspots of
  wildlife-vehicle collision on the {\c{c}}ank{\i}r{\i}-k{\i}r{\i}kkale highway
  during summer, Turkish Journal of Zoology 41~(4) (2017) 722--730.

\bibitem{roadkill2015sa}
W.~Collinson, D.~Parker, R.~Bernard, B.~Reilly, H.~Davies-Mostert, An inventory
  of vertebrate roadkill in the greater mapungubwe transfrontier conservation
  area, south africa, African Journal of Wildlife Research 45 (2015) 301--311.
\newblock \href {http://dx.doi.org/https://doi.org/10.3957/056.045.0301}
  {\path{doi:https://doi.org/10.3957/056.045.0301}}.

\bibitem{liu2020}
Y.~Liu, S.~Zhang, H.~Yu, Y.~Wang, Y.~Feng, J.~Sun, X.~Zhou, Straw segmentation
  algorithm based on modified unet in complex farmland environment, Computers,
  Materials \& Continua 66 (2020) 247--262.
\newblock \href {http://dx.doi.org/10.32604/cmc.2020.012328}
  {\path{doi:10.32604/cmc.2020.012328}}.

\bibitem{ronneberger2015u}
O.~Ronneberger, P.~Fischer, T.~Brox, U-net: Convolutional networks for
  biomedical image segmentation, in: International Conference on Medical image
  computing and computer-assisted intervention, Springer, 2015, pp. 234--241.

\bibitem{bradski2008learning}
G.~Bradski, A.~Kaehler, Learning OpenCV: Computer vision with the OpenCV
  library, " O'Reilly Media, Inc.", 2008.

\bibitem{goeldner2007tourism}
C.~R. Goeldner, J.~B. Ritchie, Tourism principles, practices, philosophies,
  John Wiley \& Sons, 2007.

\bibitem{republic1996white}
R.~of~South~Africa, White paper on the development and promotion of tourism in
  south africa (1996).

\bibitem{mansfeld2005tourism}
Y.~Mansfeld, A.~Pizam, Tourism, security and safety: From theory to practice
  (the management of hospitality and tourism enterprises).

\bibitem{wcg2022}
W.~cape government, Tourism safety strategy. destination western cape. creating
  an environment for economic growth and jobs, Tech. rep., Western cape
  government, Economic development, and tourism (2022).

\bibitem{oddone2021spatio}
A.~G. H.~E. Oddone~Aquino, S.~L. Nkomo, Spatio-temporal patterns and
  consequences of road kills: a review, Animals 11~(3) (2021) 799.

\bibitem{marcel2009}
P.~H. Marcel, D.~H. Tiffany, B.~Matt, C.~G. Mark, T.~M. Pat, H.~Barrett,
  W.~Shaowei, The comparison of animal detection systems in a test-bed: A
  quantitative comparison of system reliability and experiences with operation
  and maintenance, Tech. rep., Federal Highway Administration and Montana
  Department of Transportation (2009).

\bibitem{WilliamH2019}
W.~H.~S. Antônio, M.~D. Silva, R.~S. Miani, J.~R. Souza, A proposal of an
  animal detection system using machine learning, Applied Artificial
  Intelligence 33 (2019) 1093--1106.

\bibitem{MarcelHuijser2010}
M.~P. Huijser, L.~Hayden, Evaluation of the reliability of an animal detection
  system in a test-bed, Tech. rep., Western Transportation Institute (2010).

\bibitem{smith2016}
D.~Smith, M.~Grace, A.~Miller, M.~Noss, R.~Noss, Assessing the effectiveness
  and reliability of the roadside animal detection system on us highway 41 near
  the turner river in collier county. final report. contract no. bdv37, two
  \#2., Tech. rep., Florida Department of Transportation, District One, Bartow,
  FL. 72 pp. + appendices (2016).

\bibitem{shapoval2018}
V.~Shapoval, J.~Lev, J.~Barto{\v{s}}ka, F.~Kumh{\'a}la, Application of doppler
  radar for wildlife detection in vegetation, Scientia Agriculturae Bohemica 49
  (2018) 136--141.

\bibitem{marcel2017}
M.~P. Huijser, E.~R. Fairbank, F.~D. Abra, et~al., The reliability and
  effectiveness of a radar-based animal detection system, Tech. rep.,
  University of Alaska Fairbanks. Center for Environmentally Sustainable
  (2017).

\bibitem{vikhram2017}
B.~Vikhram, B.~Revathi, R.~Shanmugapriya, S.~Sowmiya, G.~Pragadeeswaran, Animal
  detection system in farm areas, International Journal of Advanced Research in
  Computer and Communication Engineering 6 (2017) 587--591.

\bibitem{abir2013}
A.~Mukherjee, A.~Sullivan, A.~Sinha, X.~Liu, D.~Brake, Roadway monitoring and
  driver warning systems for wildlife-vehicle collision avoidance, Tech. rep.,
  AUG Signals Ltd (2013).

\bibitem{CRISTIANDRUTA2015}
C.~Druta, A.~S. Alden, et~al., Evaluation of a buried cable roadside animal
  detection system., Tech. rep., Virginia Center for Transportation Innovation
  and Research (2015).

\bibitem{CRISTIANDRUTA2020}
C.~Druta, A.~S. Alden, Preventing animal-vehicle crashes using a smart
  detection technology and warning system, Transportation research record 2674
  (2020) 680--689.

\bibitem{sharma2017real}
S.~Sharma, D.~Shah, Real-time automatic obstacle detection and alert system for
  driver assistance on indian roads, International Journal of Vehicle
  Autonomous Systems 13~(3) (2017) 189--202.

\bibitem{rosenband2017inside}
D.~L. Rosenband, Inside waymo's self-driving car: My favorite transistors, in:
  2017 Symposium on VLSI Circuits, IEEE, 2017, pp. C20--C22.

\bibitem{sillero2018}
N.~Sillero, R.~Hélder, F.~Marc, S.~Cristiano, L.~Gil, A road mobile mapping
  device for supervised classification of amphibians on roads, European Journal
  of Wildlife Research 64.
\newblock \href {http://dx.doi.org/https://doi.org/10.1007/s10344-018-1236-4}
  {\path{doi:https://doi.org/10.1007/s10344-018-1236-4}}.

\bibitem{Diana2019}
D.~Sousa~Guedes, H.~Ribeiro, N.~Sillero, An improved mobile mapping system to
  detect road-killed amphibians and small birds, ISPRS International Journal of
  Geo-Information 8.
\newblock \href {http://dx.doi.org/https://doi.org/10.3390/ijgi8120565}
  {\path{doi:https://doi.org/10.3390/ijgi8120565}}.

\bibitem{hallisey2022estimating}
N.~Hallisey, S.~W. Buchanan, B.~D. Gerber, L.~S. Corcoran, N.~E. Karraker,
  Estimating road mortality hotspots while accounting for imperfect detection:
  A case study with amphibians and reptiles, Land 11~(5) (2022) 739.

\bibitem{nandutu2021integrating}
I.~Nandutu, M.~Atemkeng, P.~Okouma, Integrating ai ethics in wildlife
  conservation ai systems in south africa: A review, challenges, and future
  research agenda, AI \& SOCIETY (2021) 1--13.

\bibitem{gagnon2015cost}
J.~W. Gagnon, C.~D. Loberger, S.~C. Sprague, K.~S. Ogren, S.~L. Boe, R.~E.
  Schweinsburg, Cost-effective approach to reducing collisions with elk by
  fencing between existing highway structures, Human--Wildlife Interactions
  9~(2) (2015) 14.

\bibitem{ag_fd2022}
A.~Game, F.~Department, Wildlife compatible fencing,
  \url{https://www.nrcs.usda.gov/wps/PA_NRCSConsumption/download?cid=nrcseprd1080807&ext=pdf},
  Accessed 11 July 2022 (2022).

\bibitem{farm_fence2022}
J.~W. Worley, Fences for the firm,
  \url{https://www.nrcs.usda.gov/Internet/FSE_DOCUMENTS/nrcs141p2_023913.pdf},
  Accessed 11 July 2022 (2022).

\bibitem{datainbrief}
M.~Atemkeng, I.~Nandutu, P.~Okouma, A dissected dataset for single and double
  wildlife fences in south africa (under review), Data in Brief.

\bibitem{shorten2019survey}
C.~Shorten, T.~M. Khoshgoftaar, A survey on image data augmentation for deep
  learning, Journal of big data 6 (2019) 1--48.

\bibitem{dutta2019via}
A.~Dutta, A.~Zisserman, The via annotation software for images, audio and
  video, in: Proceedings of the 27th ACM international conference on
  multimedia, 2019, pp. 2276--2279.

\bibitem{albawi2017understanding}
S.~Albawi, T.~A. Mohammed, S.~Al-Zawi, Understanding of a convolutional neural
  network, in: 2017 international conference on engineering and technology
  (ICET), Ieee, 2017, pp. 1--6.

\bibitem{o2015introduction}
K.~O'Shea, R.~Nash, An introduction to convolutional neural networks, arXiv
  preprint arXiv:1511.08458.

\bibitem{saha2018transfer}
R.~Saha, Transfer learning--a comparative analysis (2018).

\bibitem{zhu2011heterogeneous}
Y.~Zhu, Y.~Chen, Z.~Lu, S.~J. Pan, G.-R. Xue, Y.~Yu, Q.~Yang, Heterogeneous
  transfer learning for image classification, in: Twenty-fifth aaai conference
  on artificial intelligence, 2011.

\bibitem{shin2016deep}
H.-C. Shin, H.~R. Roth, M.~Gao, L.~Lu, Z.~Xu, I.~Nogues, J.~Yao, D.~Mollura,
  R.~M. Summers, Deep convolutional neural networks for computer-aided
  detection: Cnn architectures, dataset characteristics and transfer learning,
  IEEE transactions on medical imaging 35~(5) (2016) 1285--1298.

\bibitem{van2014transfer}
A.~Van~Opbroek, M.~A. Ikram, M.~W. Vernooij, M.~De~Bruijne, Transfer learning
  improves supervised image segmentation across imaging protocols, IEEE
  transactions on medical imaging 34~(5) (2014) 1018--1030.

\bibitem{weiss2016survey}
K.~Weiss, T.~M. Khoshgoftaar, D.~Wang, A survey of transfer learning, vol. 3,
  no. 1 (2016).

\bibitem{he2016deep}
K.~He, X.~Zhang, S.~Ren, J.~Sun, Deep residual learning for image recognition,
  in: Proceedings of the IEEE conference on computer vision and pattern
  recognition, 2016, pp. 770--778.

\bibitem{he2016identity}
K.~He, X.~Zhang, S.~Ren, J.~Sun, Identity mappings in deep residual networks,
  in: European conference on computer vision, Springer, 2016, pp. 630--645.

\bibitem{hossin2015review}
M.~Hossin, M.~N. Sulaiman, A review on evaluation metrics for data
  classification evaluations, International journal of data mining \& knowledge
  management process 5~(2) (2015) 1.

\bibitem{li2021application}
M.-Y. Li, D.-J. Zhu, W.~Xu, Y.-J. Lin, K.-L. Yung, A.~W. Ip, Application of
  u-net with global convolution network module in computer-aided tongue
  diagnosis, Journal of Healthcare Engineering 2021.

\bibitem{zou2004statistical}
K.~H. Zou, S.~K. Warfield, A.~Bharatha, C.~M. Tempany, M.~R. Kaus, S.~J. Haker,
  W.~M. Wells~III, F.~A. Jolesz, R.~Kikinis, Statistical validation of image
  segmentation quality based on a spatial overlap index1: scientific reports,
  Academic radiology 11~(2) (2004) 178--189.

\bibitem{kingma2014adam}
D.~P. Kingma, J.~Ba, Adam: A method for stochastic optimization, arXiv preprint
  arXiv:1412.6980.

\bibitem{viola2001rapid}
P.~Viola, M.~Jones, Rapid object detection using a boosted cascade of simple
  features, in: Proceedings of the 2001 IEEE computer society conference on
  computer vision and pattern recognition. CVPR 2001, Vol.~1, Ieee, 2001, pp.
  I--I.

\bibitem{eloff2008temporal}
P.~Eloff, A.~Van~Niekerk, Temporal patterns of animal-related traffic accidents
  in the eastern cape, south africa, South African Journal of Wildlife
  Research-24-month delayed open access 38~(2) (2008) 153--162.

\bibitem{tabak2019machine}
M.~A. Tabak, M.~S. Norouzzadeh, D.~W. Wolfson, S.~J. Sweeney, K.~C.
  VerCauteren, N.~P. Snow, J.~M. Halseth, P.~A. Di~Salvo, J.~S. Lewis, M.~D.
  White, et~al., Machine learning to classify animal species in camera trap
  images: Applications in ecology, Methods in Ecology and Evolution 10~(4)
  (2019) 585--590.

\bibitem{willi2019identifying}
M.~Willi, R.~T. Pitman, A.~W. Cardoso, C.~Locke, A.~Swanson, A.~Boyer,
  M.~Veldthuis, L.~Fortson, Identifying animal species in camera trap images
  using deep learning and citizen science, Methods in Ecology and Evolution
  10~(1) (2019) 80--91.

\bibitem{liuZhenyi2020}
Z.~Liu, T.~Lian, J.~Farrell, B.~Wandell, Neural network generalization: The
  impact of camera parameters, IEEE Access PP (2020) 1--1.
\newblock \href {http://dx.doi.org/10.1109/ACCESS.2020.2965089}
  {\path{doi:10.1109/ACCESS.2020.2965089}}.

\bibitem{lecun2015deep}
Y.~LeCun, Y.~Bengio, G.~Hinton, Deep learning, nature 521~(7553) (2015)
  436--444.

\bibitem{ackermann2018deploying}
K.~Ackermann, J.~Walsh, A.~De~Un{\'a}nue, H.~Naveed, A.~Navarrete~Rivera, S.-J.
  Lee, J.~Bennett, M.~Defoe, C.~Cody, L.~Haynes, et~al., Deploying machine
  learning models for public policy: A framework, in: Proceedings of the 24th
  ACM SIGKDD International Conference on Knowledge Discovery \& Data Mining,
  2018, pp. 15--22.

\bibitem{kulkarni2021key}
V.~Kulkarni, M.~Gawali, A.~Kharat, et~al., Key technology considerations in
  developing and deploying machine learning models in clinical radiology
  practice, JMIR Medical Informatics 9~(9) (2021) e28776.

\bibitem{takabe2013rapid}
Y.~Takabe, M.~Uehara, Rapid deployment for machine learning in educational
  cloud, in: 2013 16th International Conference on Network-Based Information
  Systems, IEEE, 2013, pp. 372--376.

\bibitem{tabiaa2019deployment}
M.~Tabiaa, A.~Madani, The deployment of machine learning in ebanking: A
  survey., in: 2019 Third International Conference on Intelligent Computing in
  Data Sciences (ICDS), IEEE, 2019, pp. 1--7.

\bibitem{schonfeld2020u}
E.~Schonfeld, B.~Schiele, A.~Khoreva, A u-net based discriminator for
  generative adversarial networks, in: Proceedings of the IEEE/CVF conference
  on computer vision and pattern recognition, 2020, pp. 8207--8216.

\bibitem{wu2019u}
C.~Wu, Y.~Zou, Z.~Yang, U-gan: generative adversarial networks with u-net for
  retinal vessel segmentation, in: 2019 14th International Conference on
  Computer Science \& Education (ICCSE), IEEE, 2019, pp. 642--646.

\bibitem{prapas2021continuous}
I.~Prapas, B.~Derakhshan, A.~R. Mahdiraji, V.~Markl, Continuous training and
  deployment of deep learning models, Datenbank-Spektrum 21~(3) (2021)
  203--212.

\end{thebibliography}
\appendix

\end{document}